\documentclass[times, review, 10pt]{elsarticle}
\usepackage{lineno,hyperref}
\usepackage{array}
\usepackage{amssymb}
\usepackage{enumerate}
\usepackage{makecell}
\usepackage{mathtools}
\usepackage{array}
\usepackage{multirow}
\usepackage{ulem}
\usepackage{color}
\usepackage{diagbox}
\usepackage{enumerate}
\usepackage{graphicx}
\usepackage{threeparttable}
 \usepackage{tikz,mathpazo} 
\usepackage{indentfirst}
\usepackage{soul}
\usepackage{amsmath}
\usepackage[justification=centering]{caption}
\usepackage{tabularx}
\usepackage{algpseudocode}
\usepackage{adjustbox}
\usepackage{geometry}
\usepackage{arydshln}
\usepackage{algpseudocode}
\usepackage{hyperref}       
\usepackage{url}            
\usepackage{booktabs}       
\usepackage{amsfonts}       
\usepackage{nicefrac}       
\usepackage{microtype}      
\usepackage{xcolor}         
\usepackage{graphicx}
\usepackage{multirow}
\usepackage{makecell}
\usepackage{graphicx}
\usepackage{wrapfig}
\usepackage{caption}
\usepackage{hyperref}
\usepackage{url}
\usepackage{booktabs}       
\usepackage{amsfonts}  
\usepackage{amsmath}        
\usepackage{xcolor}         
\usepackage{boldline}
\usepackage{colortbl} 
\usepackage{subfigure}
\usepackage{subfloat}
\usepackage[ruled,noline,nofillcomment]{algorithm2e}
\usetikzlibrary{shapes.geometric, arrows}
\modulolinenumbers[5]

\soulregister{\cite}7
\soulregister{\ref}7

\newcommand{\circledsmall}[1]{\lower.7ex\hbox{\tikz\draw (0pt, 0pt)%
    circle (.5em) node {\makebox[0.1em][c]{\small#1}};}}

\newcommand{\circledtiny}[1]{\lower.7ex\hbox{\tikz\draw (0pt, 0pt)%
    circle (.3em) node {\makebox[0.1em][c]{\tiny #1}};}}

\journal{Journal of \LaTeX\ Templates}

\begin{document}

\begin{frontmatter}

\title{Adaptive Fuzzy Time Series Forecasting via Partially Asymmetric Convolution and Sub-Sliding Window Fusion}

    
    \author[label3]{Lijian Li}

    \affiliation[label3]{organization={Department of Computer and Information Science, University of Macau},
    	city={Macau},
    	postcode={999078},
    	country={China}}

\begin{abstract}
At present, state-of-the-art forecasting models are short of the ability to capture spatio-temporal dependency and synthesize global information at the stage of learning. To address this issue, in this paper, through the adaptive fuzzified construction of temporal data, we propose a novel convolutional architecture with partially asymmetric design based on the scheme of sliding window to realize accurate time series forecasting. First, the construction strategy of traditional fuzzy time series is improved to further extract short and long term temporal interrelation, which enables every time node to automatically possess corresponding global information and inner relationships among them in a restricted sliding window and the process does not require human involvement. Second, a bilateral Atrous algorithm is devised to reduce calculation demand of the proposed model without sacrificing global characteristics of elements. And it also allows the model to avoid processing redundant information. Third, after the transformation of time series, a partially asymmetric convolutional architecture is designed to more flexibly mine data features by filters in different directions on feature maps, which gives the convolutional neural network (CNN) the ability to construct sub-windows within existing sliding windows to model at a more fine-grained level. And after obtaining the time series information at different levels, the multi-scale features from different sub-windows will be sent to the corresponding network layer for time series information fusion. Compared with other competitive modern models, the proposed method achieves state-of-the-art results on most of popular time series datasets, which is fully verified by the experimental results.
\end{abstract}

\end{frontmatter}

\section{Introduction}
Time series forecasting is the task to predict future trends based on the characteristics of historical data, which is a far-reaching discipline with a very long history. Time series forecasting plays an important role in most of aspects of people's life and social operation. Accurate prediction for the future can help us make better planning, so that we can deal with various emergencies and seize important opportunities more calmly. Specifically, in the fields of electricity \cite{DBLP:journals/pr/PangZZSZ22}, environment \cite{DBLP:journals/eswa/GoliattY23}, public health \cite{DBLP:journals/tcyb/LuRN23} and transportation \cite{DBLP:journals/tai/LouLG23}, the value of practical application of time series forecasting can be realized. Nevertheless, how to effectively mine potential features and transform existing information into precise prediction is still an open issue.

Temporal data can be classified into stationary and non-stationary state. For stationary time series, lots of traditional linear models, such as ARIMA and N-BEATS, are capable of achieving quite excellent forecasting performance when the forecast horizon is relatively short. However, when it comes to long-term forecasting of non-stationary series, the situation is different, and linear models are clearly not competent for accurate prediction in more complex situations. The main reason for this circumstance is that the underlying assumptions required by linear models, such as the linearity assumption, greatly limit their ability to express nonlinear features. While real-world time series are often nonlinear, which hinders the practical application of linear models. Therefore, researchers turn their attention to nonlinear models with more powerful ability of feature expression. For instance, autoregressive conditional heteroskedasticity model \cite{articleHatemi-J} is able to accurately simulate changes in the volatility of time series variables which is a well-known and effective technique in the field of finance, therefore people can utilize it to effectively grasp market risks. Moreover, in recent years, many time series prediction methods based on deep learning have been proposed, which bring new blood to the field of time series forecasting \cite{DBLP:journals/tcyb/ChenCZWY22}. To be specific, deep learning-based prediction can be roughly divided into methods utilizing Recurrent Neural Network (RNN), Transformer and Convolutional Neural Network (CNN). 

RNN is a kind of model which receives sequence data as input and recurs in the evolution direction of the series with all cyclic units which are connected in a chain \cite{Goodfellow-et-al-2016}. Due to its special recurrent structure, the whole neural network has the ability to remember previous information, which has certain advantages in learning nonlinear characteristics of sequences. At the beginning, RNN was widely used to solve problems related to natural language processing \cite{DBLP:conf/data/FabbriM18}, and later, it was gradually extended by researchers to problems such as time series prediction \cite{DBLP:journals/tnn/IlhanKBK23} due to its efficiency in handling sequence data. However, the model based on RNN also has its restrictions. For example, it can't effectively deal with excessively long series. If the sequence exceeds a certain length, the gradient of model will disappear or explode, which also indicates that RNN is not able to remember long-term information well and taking global features into account is very tough for it. 
In this case, Transformer \cite{DBLP:conf/nips/VaswaniSPUJGKP17}, a model that is also very popular in the field of natural language processing, has entered the vision of time series researchers. Transformer solves the most common problems on RNN-based models which can make good use of GPU resources and its attention mechanism can capture long-distance feature dependencies very effectively. Researchers have utilized these features to design models that can better model information features over long distances and make more accurate predictions for both long and short-term forecasting tasks \cite{DBLP:conf/iclr/NieNSK23}. In general, both RNN-based models and methods using attention mechanism can relatively solve problems related to time series forecasting well, but there also exist their own obvious shortcomings meanwhile.

Different from the two categories of previously introduced models originating from natural language processing, CNN is the first used in the field of computer vision \cite{DBLP:conf/nips/CunBDHHHJ89}, and gradually becomes the mainstream model used in vision tasks after ResNet \cite{DBLP:conf/cvpr/HeZRS16} is proposed. The brilliance of CNN in the field of vision naturally attracts researchers of time series prediction \cite{DBLP:journals/tits/MaZDXW23}. However, the models based on CNN also have the problems that features are extracted locally, it is difficult for them to capture long-term dependencies due to lack of ability to memorize past information. The existence of these problems makes the application of CNN in time series forecasting have many obstacles. To some extent, these problems can be attributed to the fact that the receptive field of CNN is relatively limited, and the model can hardly obtain sufficient global features. Based on this consideration, researchers propose a Temporal Convolutional Network (TCN) combining causal convolution and dilated convolution \cite{DBLP:journals/corr/abs-1803-01271}. And the latter enables the time series forecasting model based on CNN to extract features at a longer distance, avoiding the problem of information loss caused by the pooling layer of traditional CNN model or higher demand of calculation resources brought by more layers of convolution. In subsequent studies, some researchers come up with a scheme to fold one-dimensional time series based on multiple periods to obtain two-dimensional tensors \cite{wu2023timesnet}, and then use modern convolutional neural network modules with large or decomposed convolution kernels for feature extraction \cite{DBLP:conf/iclr/WuHLZ0L23}. 
Generally speaking, novel CNN-based forecasting models have stronger global modeling capabilities and thus can realize relatively excellent performance in time series prediction tasks.

In this paper, we propose a novel convolutional architecture different from previous CNN-based time series forecasting models. First, we do not choose to improve the ability of capturing information over a long span of model, as TCN and similar models do. Instead, we modify the construction scheme of traditional fuzzy time series to assign location and tendency information of each element in the sliding window at the range of the whole time series data, so that the model can better obtain the global characteristics of elements in each sliding window. Moreover, the fuzzification process is carried out automatically according to the actual series conditions and there is no need to manually set parameters, which simplifies the whole construction process of fuzzification and greatly improves the reproducibility of the experiments. Second, we use the bilateral Atrous algorithm which is centered on position of original elements to initially process the allocated global information, which can greatly reduce the density of information in reconstructed sliding windows and decrease amount of calculation required by the model while retaining the location information of each element. Third, we have improved the asymmetric convolution by further decoupling it in the structure, so that filter lengths of horizontal and vertical directions can be changed accordingly for different time series datasets. To be specific, filters of disparate lengths in the horizontal and vertical directions can construct sub-windows within reconstructed windows to achieve more detailed modeling of the relationship between elements. Fourth, similar to the construction idea of many modern vision models, we utilize the scheme of using average pooling layer for down sampling and convolution with $1\times1$ kernel size for dimension change and information fusion, so as to preserve original features of input information of each layer as much as possible and to achieve a approximate effect as residual connection. More specifically, by decomposing the convolution kernel and referencing the structure of the popular vision models, the network can capture the potential connections between time nodes at different scales benefiting from the two branches of the multi-scale design.  Combining the four main changes mentioned above, the proposed model is able to achieve state-of-the-art prediction performance on most of time series datasets. To summarize briefly, the main contributions of the proposed model can be given as follows:
\begin{enumerate}
	\item By utilizing the differential temporal sequence, the proposed model applies an improved fuzzy time series construction method, which makes the model more appropriate for global information modeling by assigning corresponding fuzzy relations among remaining time nodes to each temporal elements.
	\item For further feature extraction on reconstructed time series, we use the bilateral Atrous algorithm centered on the original elements, which can reduce the amount of model calculation on the premise of retaining the location information of elements and better mine underlying temporal correlations. 
	\item By further decoupling the asymmetric convolution, partially asymmetric convolutions can construct sub-windows of different lengths in original sliding windows on separate directions, which gives the model the ability to capture potential connections between elements at a more fine-grained range, so as to provide more abundant time series features of multiple levels for prediction.
	\item Referencing the construction idea of multi-branch in visual models, average pooling and convolutional layer with kernel size $1\times1$ are utilized to store original information, so that the network can better optimize to decrease the possibility of overfitting and carry out more comprehensive feature representation.
\end{enumerate}

The rest of this paper is organized as follows. In the section of preliminary, the related theories are briefly introduced. The details of proposed method are provided in the third section. Besides, the experimental results on 43 temporal datasets are presented. Corresponding analysis of experimental results is also provided. In the last, the conclusion section is given which makes a summary of the contribution of this paper and outlook of future work.

\section{Fuzzy Time Series Forecasting}
\paragraph{Fine-grained advances in FTS (2022–2025).}
Recent work refines the earlier synopsis along three intertwined axes—model expressiveness, automated knowledge discovery, and domain-tailored hybridization.  
First, \emph{expressive neural fuzzy architectures} now dominate.  
The nine-layer SOIT2FNN-MO proposed by Yao et al.\ achieves a 25 \% lower RMSE than the classical six-layer IT2FNN on both chaotic Mackey–Glass and micro-grid datasets while retaining rule interpretability through its two-stage self-organization routine \cite{Yao2024}.  
Shafi, Lubna and Jain, Shilpi propose a novel fuzzy time series forecasting model based on hesitant fuzzy sets that integrates multiple fuzzification strategies and a weighted aggregation mechanism to effectively capture uncertainty and nondeterminism in time series data, achieving superior predictive accuracy across diverse real-world datasets.
 \cite{shafi2024improved}.  
**PhamToan, Dinh and VoThiHang** propose a novel hybrid fuzzy time series forecasting framework that integrates a self-updating clustering algorithm to dynamically construct fuzzy rules with a Bi-directional LSTM model for learning temporal dependencies, achieving state-of-the-art forecasting accuracy across benchmark datasets (M3, M4, M5) and real-world stock data (VN-index), while outperforming existing methods in both interpretability and predictive performance.
 \cite{phamtoan2024improving}.  
Zhan et al.\ address non-stationarity by differencing raw observations before a CNN-style fuzzy layer stack; their DFCNN outperforms ARIMA, Prophet, and Informer on ETTm1 with a 12 \% lower MAE and maintains stable accuracy under synthetic trend drift \cite{Zhan2023}.  

Second, \emph{data-driven partitioning and higher-order semantics} receive special attention.  
Liu \& Zhang’s IT2-FCM automatically determines both the universe of discourse and optimal rule base size, improving average coverage density index by 18 \% on Taiwan Stock Exchange data \cite{Liu2022}.  
Pinto, Arthur Caio Vargas et al. propose SODA-T2FTS, a novel univariate forecasting framework that synergistically integrates interval type-2 fuzzy logic with a self-organized, data-driven partitioning algorithm (SODA) to autonomously model temporal uncertainty and achieve state-of-the-art accuracy in complex financial time series prediction while preserving linguistic interpretability.
 \cite{pinto2022interval}.  
Ashraf, Shahzaib et al. propose a novel *q-Rung Orthopair Fuzzy Time Series (q-ROFTS)* forecasting method that extends fuzzy time series models to better handle high degrees of uncertainty and indeterminacy—particularly when traditional fuzzy set constraints are violated—by introducing a generalized, algebraically rich framework with induced q-ROFSs, max–min composition, and a MATLAB-implementable algorithm, validated through case studies on student admissions and flood forecasting \cite{ashraf2024q}.  

Third, \emph{global optimisers and hybrid domain models} now form a mature toolbox.  
Chen et al.\ blend justifiable-granularity with PSO to co-evolve partitions and rule weights, trimming parameter count by 30 \% without degrading error on NASDAQ and rainfall series \cite{Chen2025}.  
Park \& Kim’s VWFTS-PSO assigns variational weights to fuzzy relationships, raising directional-accuracy for Bitcoin, Ether, and Ripple by 5.2–6.8 \% \cite{didugu2025vwfts}.  
Wang and Lin’s sparrow-search–enhanced CEEMD-FTS lowers permutation entropy and achieves a 15 \% RMSE reduction on chaotic wind-speed data \cite{Wang2023}.  
Domain-specific hybrids flourish as well: Singh’s high-order Singh-method forecasts cut rainfall RMSE by 11 \% across 15 Indian monsoon stations \cite{Singh2024}; Chen \& Wu’s fuzzy–grey Markov chain tracks COVID-19 active-case curves with a 30 \% lower SMAPE than SEIR baselines \cite{Chen2024b}; Ahmed \& Khan embed logistic growth inside a type-1 fuzzy envelope to deliver a two-week-ahead MAPLE of 3.8 \% during the pandemic’s fourth wave \cite{Ahmed2025}.  
For finance, Kocak \& Ozkan deploy \emph{circular intuitionistic} FTS to model sector rotation, raising Sharpe ratios by 0.13 in back-tests \cite{Kocak2024}; Alam \& Wahid’s IT2-FCM stock predictor lowers RMSE by 20 \% on Bursa Malaysia blue-chips \cite{Alam2024}; finally, Liu’s PCA-driven multi-factor regressor integrates macro-economic covariates and beats VAR-GARCH by 8.9 \% MAE \cite{Liu2023}.  

Complementary developments draw from the field of uncertainty modeling and evidential reasoning. He \& Xiao\cite{he2022new, he2021conflicting} lay the foundation for conflict-aware probability assignments in fuzzy forecasting contexts. Enhancements such as MMGET\cite{he2022mmget}, TDQMF\cite{he2023tdqmf}, and ordinal belief/fuzzy entropy measures\cite{he2023ordinal, he2022ordinal} offer better quantification of indeterminacy. These techniques inspire recent evidential fusion strategies for deep learning\cite{he2024mutual, he2024uncertainty, he2025co, he2024epl}. Practical use cases emerge in temporal action localization\cite{he2024generalized}, clustering\cite{li2025adaptive}, sentiment analysis\cite{li2022nndf}, spatio-temporal vegetation tracking\cite{xu2023spatio}, and medical diagnosis with fuzzy distance matrices\cite{he2021matrix}.

Federated and multi-modal fusion research\cite{huang2025unitrans, bi2025multi, li2024efficient} further underscores the role of interpretable uncertainty in dynamic time series modeling. This is echoed in code generation~\cite{chen2025revisit}, residual inception modeling~\cite{he2024residual}, and knowledge graph completion~\cite{li2025rethinking, li2025towards, li2025multi}. Together, these works show the expanding boundary of fuzzy forecasting into broader AI systems driven by robust uncertainty-aware reasoning.

\noindent\textbf{Synthesis.}  Taken together, these 22 studies demonstrate that the field now intertwines \emph{(i)} higher-order/type-2 fuzzy semantics for uncertainty expression, \emph{(ii)} hybrid deep-learning pipelines that internalise fuzzy reasoning, and \emph{(iii)} evolutionary or meta-heuristic calibration for end-to-end optimisation, thereby delivering robust state-of-the-art accuracy across energy, finance, climate, and epidemiology forecasting tasks.

\section{Preliminaries}


\subsection{Asymmetric Convolution}Convolutional neural network (CNN) is widely applied in various fields, such as computer vision \cite{DBLP:journals/pami/0001HLC23}, speech recognition \cite{DBLP:conf/ssw/OordDZSVGKSK16} and time series forecasting \cite{DBLP:journals/tit/Liu22}. In order to improve efficiency of CNN models, researchers choose to decompose the original convolutional layer \cite{DBLP:conf/iccv/DingGDH19}. Regular convolution is redesigned into the form of individual horizontal and vertical filters, which reduces time complexity of convolution operation. Assume the original convolution as $\mathbb{C}$, and corresponding decomposed horizontal and vertical filters as $\mathbb{C}_h$ and $\mathbb{C}_v$, then the transformation of convolution can be defined as:
\begin{equation}
	\mathbb{M} \odot \mathbb{C} = \{\mathbb{M} \odot \mathbb{C}_v\} \odot \mathbb{C}_h, \mathbb{C} \in \mathbb{R}^{n\times n}, \mathbb{C}_v \in \mathbb{R}^{n\times 1}, \mathbb{C}_h \in \mathbb{R}^{1\times n}
\end{equation}
where $\odot$ is the operator of 2D convolution and $\mathbb{M}\in \mathbb{R}^{H\times W}$ is input to model. Assume $\mathbb{M^{'}}\in \mathbb{R}^{H^{'}\times W^{'}}$, due to the efficiency of asymmetric convolution, the time complexity decreases from $\mathcal{O}(n^2H^{'}W^{'})$ to $\mathcal{O}(2nH^{'}W^{'})$.

\subsection{Atrous Algorithm} Atrous algorithm is also called dilated convolution which is proposed to enlarge receptive field of convolutional neural network without increasing calculation demand \cite{DBLP:journals/corr/ChenPKMY14,DBLP:journals/pami/ChenPKMY18}. Assume there is a one-dimensional input $\mathbb{I}[i]$, its corresponding output $\mathbb{O}[i]$ and a filter $f[k]$ with $K$ length, the process of Atrous algorithm is defined as:
\begin{equation}
	\mathbb{O}[i] = \sum_{k = 1}^{K}\mathbb{I}[i + s \cdot k]f[k]
\end{equation}
where the parameter $s$ denotes the stride of dilated convolution in the process of sampling. And the convolution degenerates into the form of standard convolution when $s = 1$. Since dilated convolution can more effectively model long sequence data, some researches have also applied it to processing of time series \cite{DBLP:conf/iclr/LiuZL0L0022} to obtain global features. 

\subsection{Fuzzy Time Series}  Fuzzy mathematics \cite{DBLP:journals/iandc/Zadeh65} plays a very important role in the process of reasoning, decision making and control. Due to its effectiveness, some researchers choose to construct fuzzy relations between different time nodes \cite{DBLP:journals/tcyb/GuoWLP21}. One of concepts of traditional fuzzy time series is proposed by Chen \cite{DBLP:journals/fss/Chen96a,DBLP:journals/cas/Chen02}. Fuzzification of original temporal data gives forecasting methods the ability to model series in a relatively long range. Some subsequent fuzzy time series models \cite{DBLP:journals/tcyb/HuWCZZP22, DBLP:journals/tfs/ZhangXPDWTLL23,DBLP:journals/tcyb/HanZXQW19} combined with machine learning or deep learning algorithms have achieved excellent results in some time series prediction tasks.

\subsection{Multi-Branch Network} 

\begin{figure}
	\centering 
	\includegraphics[scale=0.3]{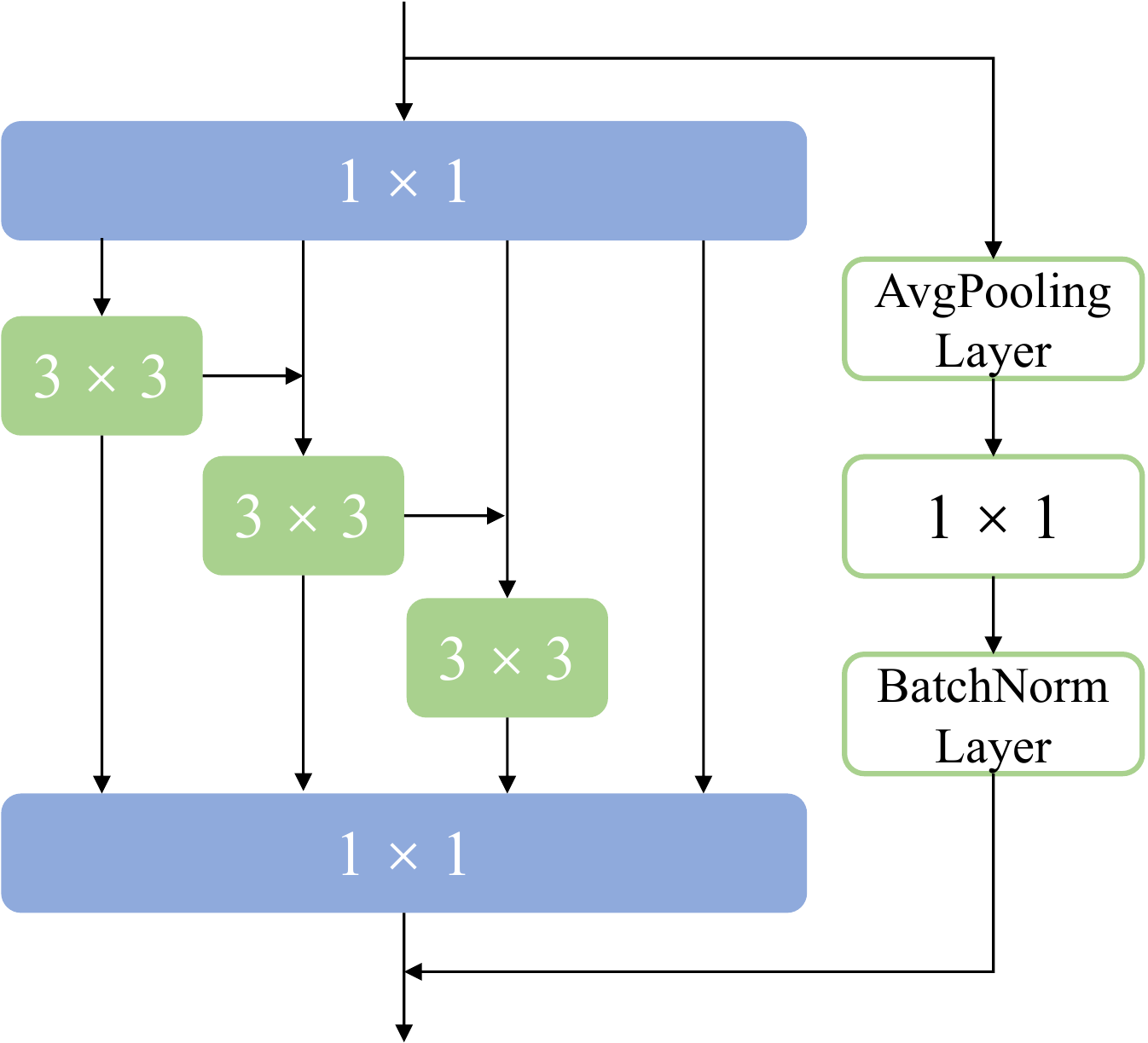}
	\caption{Details of the v1b version of Res2Net BotteleNeck}
	\label{1}
\end{figure}

Modern convolutional neural networks usually possess a multi-branch design \cite{DBLP:journals/cacm/KrizhevskySH17}. Generally speaking, when input information is processed by multiple branches, the whole network is able to receive feature representations at different levels. Specifically, inception-based networks \cite{DBLP:conf/cvpr/SzegedyVISW16, DBLP:conf/aaai/SzegedyIVA17} utilize aggregation of information from branches of network realizing relatively excellent model performance. Traditional convollutional neural network like ResNet \cite{DBLP:conf/cvpr/HeZRS16} can benefit from multi-branch design \cite{DBLP:conf/cvpr/XieGDTH17} which enables the network to acquire richer multi-level feature information. It is worth noting that there is an individual branch in some multi-branch networks to preserve more primitive characteristic information which is mainly realized by convolutional layers with $1\times1$ kernel size and pooling layers \cite{DBLP:conf/cvpr/Ding0MHD021,DBLP:journals/pami/GaoCZZYT21} in which structure of Res2Net bottleblock is given in Figure \ref{1}.

\begin{figure*}
	\centering 
	\includegraphics[scale=0.35]{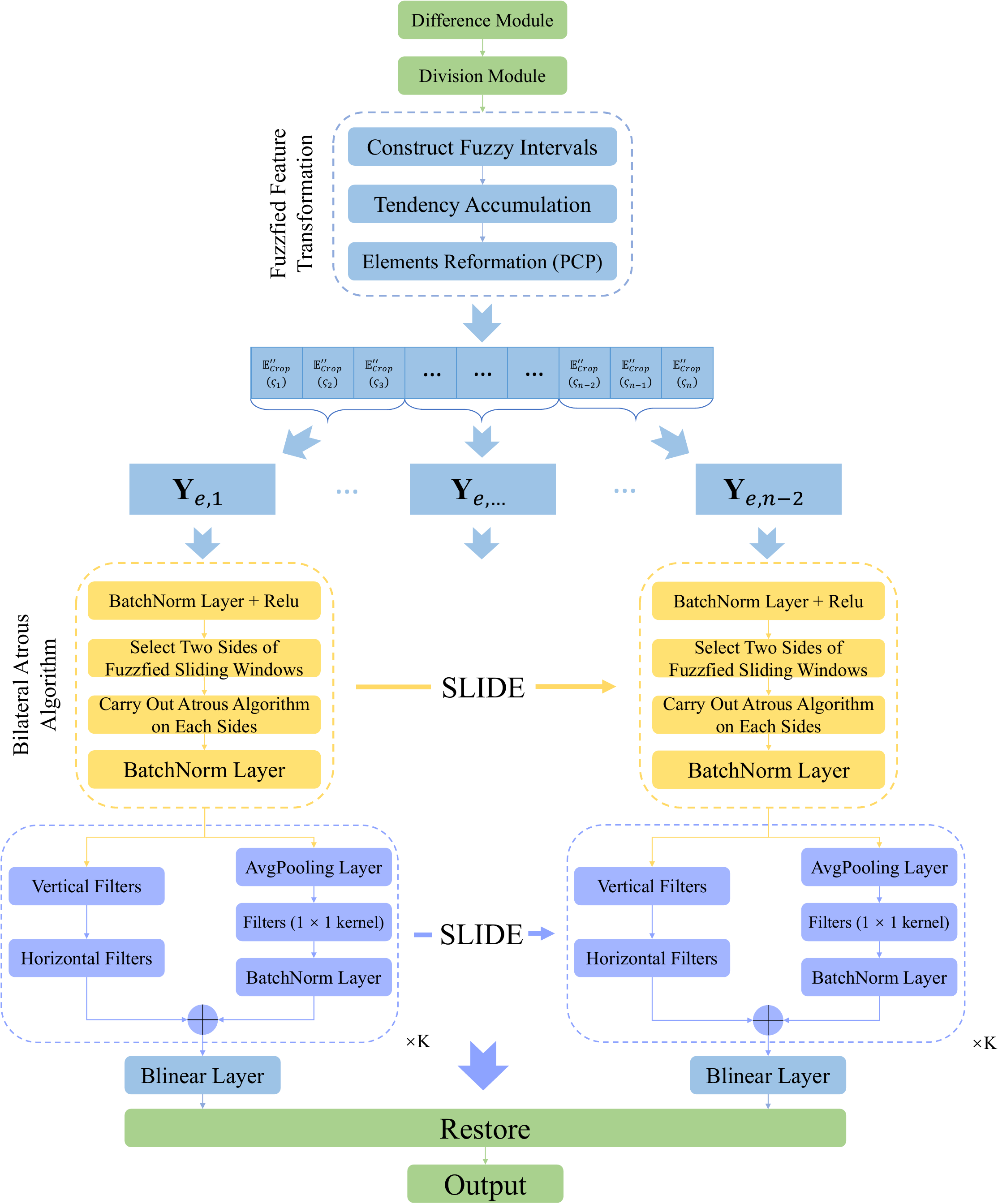}
	\caption{Details of the Proposed Model}
	\label{2}
\end{figure*}

\section{Proposed Method}
As mentioned above, fuzzification of original temporal data plays an important role in capturing long range features. Based on the concept of fuzzy time series, we mainly utilize a variant of asymmetric convolutions which possesses variable filter lengths on different directions to model correlation of time series data at disparate distances. Besides, in the stage of pre-processing and aggregation of processed information, Atrous algorithm and multi-branch network design are correspondingly improved to obtain a longer range of information acquisition and to preserve more primitive characteristic features from last network layer respectively. Besides, the detailed structure of the proposed method covering data pre-processing, fuzzified transformation, feature extraction and prediction stages is provided in Figure \ref{2} and Algorithm \ref{psuedo}.

\subsection{Difference Module}
Generally speaking, time series data usually has certain tendency and fluctuation. We choose to convert original temporal data, $\textbf{T}=\{(t_1,y_1),(t_2,y_2),...,(t_N,y_N)\}$,  into the form of difference series to make temporal data level off to some extent. Assume \textsc{Diff} is the operation of difference of series, then the process is defined as:
\begin{equation}
	\begin{split}
		\textsc{Diff}(\textbf{T}) = \{(t_{2,1}, \varsigma_1),...,(t_{N,N-1}, \varsigma_{n})\} =\\ \{(t_{2,1},y_2-y_1),...,(t_{N,N-1},y_N-y_{N-1})\}
	\end{split}
\end{equation}
where $\textsc{Diff}(\textbf{T})$ represents the difference series, $n\in[1,N-1]$ and $\varsigma_n$ represents the value of difference series at every two adjacent time points.
\subsection{Division Module}
The processed data is divided into sub-sequences according to certain settings of the sliding window. Assume the size of sliding window is $\textbf{S}$, then the partitioned series $\delta_i$ can be defined as:
\begin{equation}
	\delta_i = \{\varsigma_i, \varsigma_{i+1}, ..., \varsigma_{i+\textbf{S}-1}\}, i \in [1, n-\textbf{S} + 1]
\end{equation}

Traditional models which are designed based on the concept of sliding window only process information independently without taking global features of temporal data into consideration in the processing of data from every sliding window \cite{DBLP:journals/isci/Chu95}. The proposed method is devised to allocate each sub-sequence additional features within the corresponding complete series and reconstruct original information utilizing tendency accumulation of time points in individual sliding windows. The details are provided in the next section.

\begin{algorithm}\footnotesize
	\KwData{Original temporal data, $\textbf{T}=\{(t_1,y_1),(t_2,y_2),...,(t_N,y_N)\}$, Size of sliding window, $\textbf{S}$, predicted length $p$}
	\KwResult{Prediction $Y = \{\textbf{Y}_1,...,\textbf{Y}_p\}$}
	Convert $\textbf{T}$ into $\textsc{Diff}(\textbf{T})$;\\
	Split $\textsc{Diff}(\textbf{T})$ into sub-windows $\delta_i$;\\ 
	For i in $[1, n-\textbf{S} + 1]$:\\
	\qquad For each $\delta_i$:\\
	\qquad\qquad For $\varsigma_i$ in $\delta_i$:\\
	\qquad\qquad\qquad Place $\varsigma_i$ into corresponding interval;\\
	\qquad\qquad\qquad Calculate Tendency Accumulation of $\varsigma_i$;\\
	\qquad\qquad\qquad Obtain $\mathbb{E}^{'}(\varsigma_i)$;\\
	Calculate the total shortest length of side $SL$;\\
	For each $\mathbb{E}^{'}(\varsigma_i)$:\\
	\qquad Pad the shorter side of $\mathbb{E}^{'}(\varsigma_i)$ and obtain $\mathbb{E}^{''}(\varsigma_i)$;\\
	\qquad Crop $\mathbb{E}^{''}(\varsigma_i)$ according $SL$;\\
	\qquad Obtain $\textbf{Y}_{e,i}$;\\
	For i in $[1, n-\textbf{S} + 1]$:\\
	\qquad For each $\textbf{Y}_{e,i}$:\\
	\qquad \qquad $\textbf{Y}_{b,i} = \textsc{ReLu}(\textsc{BN}_{\gamma, \beta}(\textbf{Y}_{e,i}))$;\\
	\qquad \qquad Carry out BAA on two sides of $\textbf{Y}_{b,i}$;\\
	\qquad \qquad Obtain $\textbf{Y}_{{b,i}^{'}} $ ;\\
	\qquad \qquad Obtain $\textbf{Y}_i = \textsc{BN}(\textbf{Y}_{{b,i}^{'}})$;\\
	For i in $[1, n-\textbf{S} + 1]$:\\
	\qquad For each $\textbf{Y}_i$:\\
	\qquad\qquad Carry out partially convolution;\\
	\qquad\qquad Obtain $\textbf{Y}^{p_1}_i = [(\textbf{Y}_i*f_V)*f_H]_{\times K}$;\\
	\qquad\qquad Obtain $\textbf{Y}^{p_2}_i = \textsc{BN}(f_{1\times1} \diamond \textsc{AvgP}(\textbf{Y}_i))$;\\
	\qquad\qquad Synthesize $\textbf{Y}^{p_1}_i $ and $\textbf{Y}^{p_2}_i$;\\
	\qquad\qquad Make difference prediction $\hat{\textbf{Y}}_i$\\
	Restore $\hat{\textbf{Y}}_i$ into final prediction;\\
	
	\tcc{In the process of prediction, the prediction produced by the network will be sent to the network to satisfy the input size required by the proposed method and avoid information leak}
	\caption{The Procedure of the Proposed Method}
	\label{psuedo}
\end{algorithm}

\subsection{Fuzzified Feature Transformation Module}
According to fuzzy time series \cite{DBLP:journals/fss/Chen96a}, minimum and maximum enrollments $\epsilon_1$ and $\epsilon_2$ are subjectively selected to constitute a universe of discourse, which may lead to fluctuations in model performance and unrepeatability of experimental results. Therefore, we assign the standard deviations of the complete time series, $\sigma(\textsc{Diff}(\textbf{T}))$, to the two enrollments to avoid the problems mentioned above. Then, a universe of discourse of differential temporal data $\textsc{Diff}(\textbf{T})$ can be defined as:
\begin{equation}
	\mathbb{U}_{\textsc{Diff}(\textbf{T})} = [\varsigma_{min} - \epsilon_1, \varsigma_{max}+\epsilon_2], \epsilon_1 = \epsilon_2= \sigma(\textsc{Diff}(\textbf{T})) 
\end{equation}

where $\varsigma_{min}$ and $\varsigma_{max}$ denote the minimum and maximum element in the differential series. In order to construct different intervals in the universe of discourse $\mathbb{U}_{\textsc{Diff}(\textbf{T})}$, the number of intervals is set to $\mathbb{N} = log_2^n $. Assume the lower and upper bounds of the universe of discourse are $\alpha_l$ and $\alpha_r$, and the length of each interval is $\varphi$. For every element $\varsigma_i$ in $\delta$, it can be allocated a certain position in the universe of discourse. With respect to the element $\varsigma_{i+\textbf{S}-1}$, it can be expanded into the form:
\begin{equation}
	\begin{split}
		\mathbb{E}(\varsigma_{i+\textbf{S}-1}) = [\alpha_l, ..., \alpha_{l} + \varphi * \tau, \varsigma_{i+\textbf{S}-1}, \alpha_{r} +\\ (\varphi+1) * \tau, ..., \alpha_r]^\top, \varphi\in[1,\mathbb{N}], \tau = \frac{\alpha_r - \alpha_l}{\mathbb{N}}
	\end{split}
\end{equation}
where $\mathbb{E}(\varsigma_{i+\textbf{S}-1})$ represents the operation of expansion of every element in $\delta_i$. Besides, to reflect the changing trend of data before each time point, we defined a variable called tendency accumulation $\varrho$. For the element $\varsigma_{i+\textbf{S}-1}$, it can be defined as:
\begin{equation}
	\varrho = \sum_{j = i}^{i+\textbf{S}-2} (\frac{(\varsigma_{i+\textbf{S}-1} - \varsigma_j)(t_{i+\textbf{S}} - t_{i+\textbf{S}-1})}{t_{i+\textbf{S}-1} - t_j}) / (\textbf{S}-1)
\end{equation}

Specifically, the reconstructed form of  $\varsigma_{i+\textbf{S}-1}$ is $\varsigma_{i+\textbf{S}-1} + \varrho$. And correspondingly, $\mathbb{E}(\varsigma_{i+\textbf{S}-1})$ can be rewritten as:
\begin{equation}
	\mathbb{E}^{'}(\varsigma_{i+\textbf{S}-1}) = [\alpha_l, ..., \alpha_{l} + \varphi * \tau, \varsigma_{i+\textbf{S}-1}+ \varrho, \alpha_{r} + (\varphi+1) * \tau, ..., \alpha_r]^\top
\end{equation}

Every element in $\delta_i$ is reformed as above. And a notable point is that each reconstructed form of element $\varsigma_i$ in $\delta_i$ is not exactly in the same position which may lead to difficulty of following processing of information. So, we design a padding-crop policy (PCP) to make sure the elements are located in the same position. Let's take $\varsigma_{i+\textbf{S}-1}$ as an example again. Specifically, we first pad the shorter side of $E^{'}(\varsigma_{i+\textbf{S}-1})$ with the same value $\alpha_l$ or $\alpha_r$ which depends on the direction of shorter side. If the right side is shorter, the process can be defined as:
\begin{equation}
	\begin{split}
		\mathbb{E}^{''}(\varsigma_{i+\textbf{S}-1}) = [\alpha_l, ..., \alpha_{l} + \varphi * \tau, \varsigma_{i+\textbf{S}-1}+ \varrho, \alpha_{r} + (\varphi+1) * \tau, ..., \\ \widehat{\alpha_r,...,\alpha_r}]^\top, \\Length(\alpha_l\rightarrow\alpha_{l} + \varphi * \tau) = Length(\alpha_{l} + (\varphi+1) * \tau\rightarrow\widehat{\alpha_r\rightarrow\alpha_r})
	\end{split}
\end{equation}
where $\widehat{\alpha_r,...,\alpha_r}$ denotes a vector consist of the same element $\alpha_r$ and $Length(a,b)$ represents the length of vector whose element ranges from $a$ and $b$. Then, we conduct the same operation to every element $\varsigma_{i}$ in $\delta_i$. After padding, it is easy to find the shortest length $SL$ of two sides of the elements. And we crop every padded vector according to the shortest length from ends of every two sides of $\mathbb{E}^{''}(\varsigma_{i})$ to make sure the lengths of them are identically the same obtaining $\mathbb{E}^{''}_{Crop}(\varsigma_{i})$.

\begin{table}\scriptsize
	\renewcommand{\arraystretch}{0.7}
	\caption{Abbreviations of Datasets}
	\centering
	\begin{tabular}{cccc}
		\toprule 
		Datasets & Abbreviations  &  Datasets & Abbreviations\\ 
		\midrule 
		M1 Yearly & M1 Y & M1 Quarterly & M1 Q \\
		M1 Monthly & M1 M & M3 Yearly & M3 Y \\
		M3 Quarterly & M3 Q & M3 Other & M3 O \\
		M4 Yearly& M4 Y & M4 Monthly & M4 M \\
		M4 Weekly & M1 Y & M4 Daily & M4 D \\
		M4 Hourly & M1 Y & Tourism Yearly & TY \\
		Tourism Quarterly & TQ & Tourism Monthly & TM \\
		Aus. Electricity Demand & A.E.D & Pedestrian Counts & PC\\
		Vehicle Trips & VT & KDD Cup & KDD\\
		Saugeen River Flow & SRF & NN5 Daily & NN5 D\\
		NN5 Weekly & NN5 W & Web Traffic Daily & WTD\\
		Web Traffic Weekly & WTW & Solar 10 Minutes & S10\\
		Solar Weekly & SW & Electricity Hourly & EH\\
		Electricity Weekly & EW & FRED-MD & F-M\\
		Traffic Hourly & TH & Traffic Weekly & TW\\
		COVID Deaths & COVID & Temperature Rain & TR\\
		\bottomrule 
	\end{tabular}
	\label{abbreviation}
\end{table}	

\begin{table*} \footnotesize
	\renewcommand{\arraystretch}{0.6}
	\caption{MEAN MAE RESULTS OF SINGLE-VARIATE DATASETS}
	\label{table2}
	\centering
	\setlength{\tabcolsep}{1.1mm}{
		\begin{tabular}{|l| c c c c c c c c c c|}
			\toprule
			Dataset & SES & Theta & TBATS & ETS & ARIMA & PR& CatBoost &  FFNN & DeepAR &Informer\\
			\midrule
			TQ &  15014.19 &7656.49&9972.42 &8925.52 &10475.47 &9092.58 &10267.97& 8981.04 &9511.37&-\\
			TM& 5302.10 & 2069.96 & 2940.08& 2004.51 & 2536.77 & 2187.28 & 2537.04&  2022.21 & \textcolor{red}{\textbf{1871.69}}&-\\
			TY &  95579.23 &90653.60 &94121.08 &94818.89 &95033.24 &82682.97 &79567.22 &79593.22 &71471.29&-\\
			A.E.D &  659.60 &665.04 &370.74 &1282.99 &1045.92 &247.18 &241.77& 258.76 &302.41&-\\
			Bitcoin & 5.33$\times$10$^{18}$ & 5.33$\times$10$^{18}$ & 9.9$\times$10$^{17}$ & 1.1$\times$10$^{18}$ & 3.62$\times$10$^{18}$ & 6.66$\times$10$^{17}$ & 1.93$\times$10$^{18}$ & 1.45$\times$10$^{18}$ & 1.95$\times$10$^{18}$&-\\
			CIF 2016 & 581875.97 &714818.58 &855578.40 &642421.42 &469059.49 &563205.57& 603551.30 &1495923.44 &3200418.00&-\\
			Dominick & 5.70 & 5.86 & 7.08 & 5.81 & 7.10 & 8.19 & 8.09 & 5.85 & 5.23&-\\
			KDD&  42.04 &42.06 &39.20 &44.88 &52.20 &36.85 &34.82 &37.16 &48.98&-\\
			PC& 170.87 &170.94 &222.38 &216.50&635.16 &44.18 &\textcolor{red}{\textbf{43.41} }&46.41 &44.78-&\\
			VT& 29.98 &30.76 &21.21 &30.95 &30.07 &27.24 &22.61 &22.93 &22.00&-\\
			Weather &  2.24 &2.51 &2.30 &2.35 &2.45 &8.17 &2.51 &2.09 &2.02&-\\
			SRF &  21.50 &21.49 &22.26 &30.69 &22.38 &25.24& 21.28 &22.98 &23.51&28.59\\
			Sunspot &4.93&4.93 &2.57 &4.93 &2.57& 3.83 &2.27 &7.97 &0.77&19.43\\
			US Births& 1192.20 &586.93 &\textcolor{red}{\textbf{399.00}}&419.73 &526.33 &574.93 &441.70 &557.87 &424.93&609.43\\
			M1 Q &  2206.27 & 1981.96 &  2326.46 & 2088.15 & 2191.10 &  1630.38 & 1802.18 & 1617.39 &1951.14&-\\
			M1 M & 2259.04 & 2166.18 & 2237.50 & 1905.28 &  2080.13 &  2088.25 & 2052.32 & 2162.58&1860.81&-\\
			M1 Y &   171353.41 &  152799.26 & 103006.90 & 146110.11 & 145608.87 &   134246.38 &  215904.20 & 136238.80 & 152084.40&-\\
			M3 Q & 571.96 & 486.31 & 561.77 & 513.06 &  559.40 &  519.30 & 593.29 &  528.47 & 519.35&-\\
			M3 M & 743.41 & 623.71 & 630.59 &  626.46 & 654.80 &  692.97 & 732.00 & 692.48 &728.81&-\\
			M3 Y &  1022.27 & 957.40 & 1192.85 & 1031.40 & 1416.31 &  1018.48 &  1163.36 &  1082.03 & 994.72&-\\
			M3 O & 277.83 &  215.35 & 189.42 & 194.98 & 193.02 & 234.43 &  318.13 & 240.17 & 247.56&-\\
			M4 Q &  622.57 &574.34& 570.26 &573.19 &604.51 &610.51 &609.55 &631.01 &597.16&-\\
			M4 M & 625.24 &563.58 &589.52 &582.60 &575.36 &596.19 &611.69 &612.52 &615.22&-\\
			M4 Y & 1009.06 &  890.51 & 960.45 &  920.66 & 1067.16 & 875.76 & 929.06 & - &-&-\\
			M4 D & 178.27 &178.86 &176.60 &193.26 &179.67 &181.92 &231.36 &177.91 &299.79&-\\
			M4 H &  1218.06 &1220.97 &386.27 &3358.10 &1310.85 &257.39 &285.35 &385.49 &886.02&-\\
			M4 W & 336.82 &333.32 &296.15 &335.66 &321.61 &293.21 &364.65 &338.37 &351.78&-\\
			\midrule
			Dataset  & N-BEATS & WaveNet & Transformer& Prophet & MSS$^{\bullet}$& FEDformer$^{\bullet}$& NetAtt$^{\bullet}$&  Pyraformer$^{\bullet}$  & PFSD$^{\bullet}$ & Ours\\
			\midrule
			TQ & 8640.56 & 9137.12 & 9521.67 & 13312.35& - & - & - & - & -& \textcolor{red}{\textbf{5616.26}}\\
			TM& 2003.02 & 2095.13 & 2146.98&3442.83 & -& - & - & -&  -& 2553.89\\
			TY &  70951.80 & 69905.47 & 74316.52 &86011.16	& - & -& -& - &- & \textcolor{red}{\textbf{54038.14}} \\
			A.E.D & 213.83 & 227.50 & 231.45 &325.47& - & - & - & - & - & \textcolor{red}{\textbf{39.30}} \\
			Bitcoin &  1.06$\times$10$^{18}$ & 2.46$\times$10$^{18}$ & 2.61$\times$10$^{18}$&3.86$\times$10$^{18}$& - & - & - & - & - &\textcolor{red}{\textbf{3.77$\times$10$^{17}$}} \\
			CIF 2016 & 679034.80 & 5998224.62& 4057973.04&1053054.23 & - & - & - & - & - & \textcolor{red}{\textbf{202536.28}}\\
			Dominick &  8.28 & 5.10 & 5.18&7.85& 5.39 & 5.10 & 6.02 & 5.16 & 4.80&\textcolor{red}{\textbf{4.28}} \\
			KDD &  49.10 & 37.08 &  44.46 &34.23& - & - & - & - & - & \textcolor{red}{\textbf{5.75}}\\
			PC &  66.84 & 46.46 & 47.29 &345.33& - & - & - & - & - & 53.76 \\
			VT & 28.16 & 24.15 & 28.01 &32.61& -& - & - & - & - &\textcolor{red}{\textbf{11.20}} \\
			Weather & 2.34 & 2.29 & 2.03&2.87 & - & - & -& - & - & \textcolor{red}{\textbf{1.73}}\\
			SRF &  27.92 & 22.17 & 28.06&22.77& - & -& - & - & - & \textcolor{red}{\textbf{8.62}} \\
			Sunspot &  14.47 &  0.17& \textcolor{red}{\textbf{0.13}} &32.87& -& - & - & - & -& 0.14 \\
			US Births&  422.00&  504.40&  452.87 &1544.60& - & - & -& -& -& 469.14\\
			M1 Q &  1820.25 & 1855.89&  1864.08 &1656.51& 1686.22&1683.57 & 1727.60 & 1721.32 & 1231.13&  \textcolor{red}{\textbf{1203.76}} \\ 
			M1 M & 1820.37 & 2184.42& 2723.88&1656.51 &2063.19& 2394.66& 1720.12& 2421.01 & 1952.81& \textcolor{red}{\textbf{1540.12}}\\
			M1 Y&   173300.20 &   284953.90 & 164637.90&167627.15 & 59228.64 & 124729.30 &  66409.64 & 127110.48 & \textcolor{red}{\textbf{51417.35}}& 66024.18\\
			M3 Q & 494.85 & 523.04& 719.62&653.44 & 538.85&623.58 & 591.25 & 711.46 &473.84&  \textcolor{red}{\textbf{294.51}} \\
			M3 M & 648.60 & 699.30 &  798.38&928.43 &1127.37& 728.60 & 1014.96 & 693.24 & 912.28& \textcolor{red}{\textbf{503.92}}\\
			M3 Y &  962.33 &  987.28 & 924.47&1281.34 & 933.80&873.74 & 906.63 & 891.88 & 858.70 & \textcolor{red}{\textbf{533.85}}\\
			M3 O & 221.85&  245.29 &  239.24&397.68 & 229.01&217.03 & 297.44& 196.81 &210.80& \textcolor{red}{\textbf{92.04}}\\
			M4 Q & 580.44 &  596.78 &  637.60&773.47& 560.72 & 594.24 & 617.30 & 608.55 &445.09& \textcolor{red}{\textbf{335.76}}\\
			M4 M & 578.48 &  655.51 & 780.47&732.09 & 644.51 & 688.95 & 781.42& 694.29 & 608.31& \textcolor{red}{\textbf{329.97}}\\
			M4 Y & - & - & - &1233.35& 792.87 & 730.24 & 967.37& 757.92 &528.36& \textcolor{red}{\textbf{406.09}}\\
			M4 D &  190.44 & 189.47 &  201.08&612.67 & 173.20 & 167.05& 207.44& 161.36 & 103.28& \textcolor{red}{\textbf{61.01}}\\
			M4 H &  425.75 &  393.63 & 320.54&341.31 & 1355.21 & 246.33 & 1841.90& 228.87 & 999.83& \textcolor{red}{\textbf{84.10}}\\
			M4 W & 277.73 & 359.46 & 378.89&680.54 & 301.26 & 317.16 & 322.59& 295.60&250.68&\textcolor{red}{\textbf{213.73}}\\
			\bottomrule
	\end{tabular}}
\end{table*}

\begin{table*} \footnotesize
	\renewcommand{\arraystretch}{0.6}
	\caption{MEAN RMSE RESULTS OF SINGLE-VARIATE DATASETS}
	\label{table3}
	\centering
	\setlength{\tabcolsep}{1.1mm}{
		\begin{tabular}{|l| c c c c c c c c c c  |}
			\toprule
			Dataset & SES & Theta & TBATS & ETS & ARIMA & PR& CatBoost &  FFNN & DeepAR& Informer\\
			\midrule
			TQ & 17270.57&9254.63&12001.48&10812.34&12564.77&11746.85&12787.97&12182.57&11761.96&-\\
			TM&7039.35 &2701.96&3661.51&\textcolor{red}{\textbf{2542.96}}&3132.40&2739.43&3102.76&2584.10&2359.87&-\\
			TY & 106665.20 &99914.21&105799.40&104700.51&106082.60&89645.61&87489.00&87931.79&78470.68&-\\
			A.E.D &  766.27&771.51&446.59&1404.02&1234.76&319.98&300.55&330.91&357.00&-\\
			Bitcoin& 5.35$\times$10$^{18}$&5.35$\times$10$^{18}$&1.16$\times$10$^{18}$&1.22$\times$10$^{18}$&3.96$\times$10$^{18}$&8.29$\times$10$^{18}$&2.02$\times$10$^{18}$&1.57$\times$10$^{18}$&2.02$\times$10$^{18}$&-\\
			CIF 2016 &657112.42 &804654.19&940099.90&722397.37&526395.02&648890.31&705273.30&1629741.53&3532475.00&-\\
			Dominick & 6.48&6.74&8.03&6.59&7.96&9.44&9.15&6.79&6.67&-\\
			KDD &  73.81&73.83&71.21&76.71&82.66&68.20&65.71&68.43&80.19&-\\
			PC & 228.14&228.20&261.25&278.26&820.28&61.84&\textcolor{red}{\textbf{60.78}}&67.17&65.77&-\\
			VT &36.53 &37.44&25.69&37.61&34.95&31.69&27.28&27.88&26.46&-\\
			Weather &  2.85&3.27&2.89&2.96&3.07&9.08&3.09&2.81&2.74&-\\
			SRF &  39.79&39.79&42.58&50.39&43.23&47.70&39.32&40.64&45.28&44.42\\
			Sunspot &4.95&4.95&2.97&4.95&2.96&3.95&2.38&8.43&1.14&20.31\\
			US Births& 1369.50&735.51&\textcolor{red}{\textbf{606.54}}&607.20&705.51&732.09&618.38&726.72&683.99&734.44\\
			M1 Q &2545.73 &2282.65&2673.91&2408.47&2538.45&1909.31&2161.01&1871.85&2313.32&- \\
			M1 Y& 193829.49  &171458.07&116850.90&167739.02&175343.75&152038.68&237644.50&154309.80&173075.10&-\\
			M1 M &2725.83 &2564.88&2594.48&2263.96&2450.61&2478.88&2461.68&2527.03&2202.19&-\\
			M3 Q &670.56 &567.70&653.61&598.73&650.76&605.50&697.96&621.73&606.56&-\\
			M3 M &893.88 &753.99&765.20&755.26&790.76&830.04&874.20&833.15&873.71&-\\
			M3 Y &1172.85 &1106.05&1386.33&1189.21&1662.17&1181.81&1341.70&1256.21&1157.88& -\\
			M3 O &309.68 &242.13&216.95&224.08&220.77&262.31&349.90&268.99&277.74&-\\
			M4 Q &732.82  &673.15&672.74&674.27&709.99&711.93&714.21&735.84&700.32&-\\
			M4 M & 755.45&683.72&743.41&705.70&702.06&720.46&734.79&743.47&740.26&-\\
			M4 Y &1154.49 &1020.48&1099.95&1052.12&1230.35&1000.18&1065.02&-&-&-\\
			M4 D &209.75 &210.37&208.36&229.97&212.64&213.01&263.13&209.44&343.48&-\\
			M4 H &  1476.81&1483.70&469.87&3830.44&1563.05&312.98&344.62&467.89&1095.10&-\\
			M4 W & 412.60&405.17&356.74&408.50&386.30&350.29&420.84&399.10&422.18&-\\
			\midrule
			Dataset  & N-BEATS & WaveNet & Transformer&Prophet & MSS$^{\bullet}$& FEDformer$^{\bullet}$& NetAtt$^{\bullet}$&  Pyraformer$^{\bullet}$  & PFSD$^{\bullet}$& Ours\\
			\midrule
			TQ & 11305.95&11546.58&11724.14& 14814.80&-&-&-&-&-&\textcolor{red}{\textbf{7265.60}} \\
			TM&  2596.21&2694.22&2660.06&  3895.56&-&-&-&-&-&3497.11\\
			TY &78241.67 &77581.31&80089.25& 91873.97&-&-&-&-&-&\textcolor{red}{\textbf{64762.87}}\\
			A.E.D &  268.37&286.48&295.22& 409.25&-&-&-&-&-&\textcolor{red}{\textbf{58.82}}\\
			Bitcoin & 1.26$\times$10$^{18}$&2.55$\times$10$^{18}$&2.67$\times$10$^{18}$&3.90$\times$10$^{18}$&&&&&& \textcolor{red}{\textbf{4.39$\times$10$^{17}$}}\\
			CIF 2016 & 772924.30 &6085242.41&4625974.00& 1232225.35&-&-&-&-&-&\textcolor{red}{\textbf{265326.09}}\\
			Dominick &   9.78&6.81&6.63&8.88&7.39&6.97&7.02&6.89&6.56&\textcolor{red}{\textbf{6.54}}\\
			KDD & 80.39&68.87&76.21& 64.17&-&-&-&-&-&\textcolor{red}{\textbf{9.94}}\\
			PC &  99.33&67.99&70.17&417.67&-&-&-&-&-&75.85 \\
			VT &  33.56&28.99&32.98& 38.79&-&-&-&-&-&\textcolor{red}{\textbf{14.78}}\\
			Weather & 3.09&2.98&2.81&3.53&-&-&-&-&-&\textcolor{red}{\textbf{2.50}}\\
			SRF &  48.91&42.99&49.12&42.99&-&-&-&-&-&\textcolor{red}{\textbf{18.26}}\\
			Sunspot &  14.52 &0.66&0.52&32.92&-&-&-&-&-&\textcolor{red}{\textbf{0.51}}\\
			US Births& 627.74&768.81&686.51& 1796.94&-&-&-&-&-&713.33 \\
			M1 Q &  2267.27 &2271.68&2231.50&1903.80&1977.00&1992.56&2057.60&2026.49&1458.75& \textcolor{red}{\textbf{1428.36}}\\
			M1 M & 2183.37&2578.93&3129.84& 2938.63&2427.46&2918.05&\textcolor{red}{\textbf{2024.08}}&2957.84&2369.96& 2036.42\\
			M1 Y&  192489.80&312821.80&182850.60& 177710.58&68119.81&143607.73&81092.33&145991.89&\textcolor{red}{\textbf{59867.94}}&81291.28\\
			M3 Q &  582.83&606.75&819.18& 724.78&636.68&735.21&693.52&810.20&568.22& \textcolor{red}{\textbf{365.70}}\\
			M3 M & 796.91&845.30&948.40&1091.86&1311.49&877.76&1193.29&836.84&1079.11& \textcolor{red}{\textbf{654.39}}\\
			M3 Y & 1117.37 &1147.62&1084.75&1400.83&1079.09&1019.83&1061.72&1054.66&981.94&\textcolor{red}{\textbf{662.56}} \\
			M3 O &  248.53&276.97&271.02& 417.29&260.48&245.08&335.88&227.20&247.66& \textcolor{red}{\textbf{114.94}}\\
			M4 Q &  684.65&696.96&739.06&860.25&662.18&691.68&715.13&712.44&514.54&\textcolor{red}{\textbf{423.08}} \\
			M4 M & 705.21&787.94&902.38&857.20&778.20&831.55&902.91&853.13&720.67& \textcolor{red}{\textbf{442.99}}\\
			M4 Y &  -&-&-&1314.52&898.74&787.35&1173.95&816.41&606.06&\textcolor{red}{\textbf{508.68}} \\
			M4 D   &221.69&220.45&233.63& 634.33&205.22&192.67&249.70&183.30&118.88&\textcolor{red}{\textbf{83.33}}\\
			M4 H &  501.19&468.09&391.22&417.12&1643.46&304.69&2124.99&284.09&1209.48&\textcolor{red}{\textbf{124.75}}\\
			M4 W & 330.78&437.26&456.90& 754.74&354.97&379.04&388.03&337.62&320.38& \textcolor{red}{\textbf{277.45}}\\
			\bottomrule
	\end{tabular}}
\end{table*}

		\begin{table*}
			\renewcommand{\arraystretch}{0.85}
			\caption{MEAN MAE RESULTS OF MULTI-VARIATE DATASETS}
			\label{table4}
			\centering
			\setlength{\tabcolsep}{0.3mm}{
				\begin{tabular}{|l| c c c c c c c c c c|}
					\toprule
					Dataset & SES & Theta & TBATS & ETS & ARIMA & PR& CatBoost &  FFNN & DeepAR & Informer\\
					\midrule
					COVID &353.71 &321.32 &96.29 &85.59 &85.77 &347.98 & 475.15&144.14 &201.98&-\\
					Carparts &0.55 &0.53 & 0.58& 0.56& 0.56&0.41 &0.53 &\textcolor{red}{\textbf{0.39}} &\textcolor{red}{\textbf{0.39}}&-\\
					F-M& 2798.22&3492.84 & 1989.97& 2041.42& 2957.11& 8921.94&2475.68 &2339.57 &4263.36&32700.73\\
					TW & 1.12& 1.13& 1.17& 1.14&1.22 &1.13 &1.17 &1.15 &1.18&1.42\\
					TH &0.03 &0.03 &0.04 &0.03 &0.04 &0.02 & 0.02& 0.01&0.01&0.02\\
					Rideshare &6.29 &7.62 &6.45 &6.29 &3.37 &6.30 &6.07 &6.59 &6.28&-\\
					Hospital & 21.76& 18.54& 17.43& 17.97& 19.60& 19.24& 19.17& 22.86&18.25&38.32\\
					TR & 8.18&8.22 &7.14 & 8.21& 7.19& 6.13& 6.76&5.56 &5.37&-\\
					NN5 W &15.66 & 15.30& 14.98& 15.70& 15.38&14.94 &15.29 &15.02 &14.69&19.45\\
					NN5 D &6.63 &3.80 & \textcolor{red}{\textbf{3.70}}& 3.72& 4.41&5.47 & 4.22& 4.06&3.94&4.07\\
					WTW & 2337.11&2373.98 &2241.84 & 2668.28& 3115.03& 4051.75&10715.36 & 2025.23&2272.58&-\\
					WTD &363.43 & 358.73& 415.40& 403.23& 340.36&- &- &- &-&-\\
					S10 & 3.28& 3.29& 8.77&3.28 & 2.37& 3.28&5.69 &3.28 &3.28&3.67\\
					SW & 1202.39& 1210.83& 908.65& 1131.01& 839.88&1044.98 & 1513.49& 1050.84&721.59&2360.71\\
					EW &74149.18 & 74111.14& 24347.24& 67737.82& 28457.18&44882.52 &34518.43 &27451.83 &50312.05&47773.67\\
					EH & 845.97& 846.03&574.30 &1344.61 & 868.20& 537.38& 407.14&354.39 &329.75&441.77\\
					\midrule
					Dataset  & N-BEATS & WaveNet & Transformer&Prophet & MSS$^{\bullet}$& FEDformer$^{\bullet}$& NetAtt$^{\bullet}$&  Pyraformer$^{\bullet}$  & PFSD$^{\bullet}$ & Ours\\
					\midrule
					COVID & 158.81&1049.48 &408.66 & 648.97& -& -&-&- &- & \textcolor{red}{\textbf{7.50}}\\
					Carparts & 0.98&0.40 &\textcolor{red}{\textbf{0.39}}& 0.56 & -&- &- &- &- &0.49 \\
					F-M& 2557.80&2508.40 & 4666.04&12651.79& -& -& -& -&-& \textcolor{red}{\textbf{583.61}}\\
					TW & 1.11& 1.20&1.42 & 1.18& -&- &-& -&- & \textcolor{red}{\textbf{1.09}}\\
					TH & 0.02& 0.02& 0.01&0.02&- &- &-& -&- & \textcolor{red}{\textbf{0.008}}\\
					Rideshare & 5.55& 2.75& 6.29&7.40& -& -& -& -&-&\textcolor{red}{\textbf{0.79}} \\
					Hospital & 20.18& 19.35& 36.19& 17.07& -&- &- & -&- &\textcolor{red}{\textbf{16.11}}\\
					TR &7.28 & 5.81 & 5.24&7.06& -&-& -& -&- & \textcolor{red}{\textbf{4.53}}\\
					NN5 W &\textcolor{red}{\textbf{14.19}} &19.34 & 20.34& 15.76&- &- &- &- &- &14.87 \\
					NN5 D & 4.92&3.97 &4.16 &3.78&- &- &- &- &- & 4.27\\
					WTW & 2051.30& 2025.50& 3100.32&4765.12&- &- &- &- &- & \textcolor{red}{\textbf{1437.52}}\\
					WTD & -& -& -&875.50& -& -& -& -& -& \textcolor{red}{\textbf{217.46}}\\
					S10 & 3.52&- &3.28 &3.36& 4.14&3.18 &3.93&3.22 &2.17 & \textcolor{red}{\textbf{0.07}} \\
					SW & 1172.64& 1996.89& 576.35&1471.27& 841.69& \textcolor{red}{\textbf{479.30}}& 1247.77& 513.24&649.22& 674.19\\
					EW & 32991.72& 61429.32& 76382.47& 31098.93& -& -& -& -&- &\textcolor{red}{\textbf{16152.51}} \\
					EH & 350.37&286.56 &398.80 & 462.72& -& -& -& -& -& \textcolor{red}{\textbf{202.55}}\\
					\bottomrule
			\end{tabular}}
		\end{table*}

		\begin{table*}
			\renewcommand{\arraystretch}{0.85}
			\caption{MEAN RMSE RESULTS OF MULTI-VARIATE DATASETS}
			\label{table5}
			\centering
			\setlength{\tabcolsep}{0.3mm}{
				\begin{tabular}{|l| c c c c c c c c c c|}
					\toprule
					Dataset & SES & Theta & TBATS & ETS & ARIMA & PR& CatBoost &  FFNN & DeepAR & Informer\\
					\midrule
					COVID & 403.41& 370.14& 113.00& 102.08& 100.46& 394.07& 607.92& 173.14&230.47&-\\
					Carparts &0.78 &0.78 &0.84 & 0.80& 0.81& \textcolor{red}{\textbf{0.73}}& 0.79&0.74 &0.74&-\\
					F-M& 3103.00& 3898.72& 2295.74& 2341.72&3312.46 &9736.93 &2679.38 &2631.4 &4638.71&32867.61\\
					TW &1.51 &1.53 &1.53 &1.53 &1.54 &1.50 &1.50 &1.55 &1.51&1.76\\
					TH & 0.04& 0.04& 0.05& 0.04& 0.04& 0.03&0.03 &0.02 &0.02&0.04\\
					Rideshare & 7.17& 8.60& 7.35&7.17 &4.80 & 7.18& 6.95& 7.14&7.15&-\\
					Hospital & 26.55& 22.59& 21.28& 22.02& 23.68&23.48 &23.45 &27.77 &22.01&44.25\\
					TR & 10.34& 10.36& 9.20&10.38 &9.22 &9.83 &8.71 &8.89 &9.11&-\\
					NN5 W & 18.82& 18.65& 18.53& 18.82& 18.55& 18.62& 18.67& 18.29&18.53&23.03\\
					NN5 D & 8.23& 5.28& \textcolor{red}{\textbf{5.20}}& 5.22& 6.05& 7.26& 5.73& 5.79&5.50&5.52\\
					WTW & 2970.78& 3012.39& 2951.87& 3369.64& 3777.28& 4750.26& 14040.64&2719.65 &2981.91&-\\
					WTD & 590.11&583.32 & 740.74& 650.43& 595.43& -& -& -&-&-\\
					S10& 7.23&7.23 &10.71 & 7.23& 5.55& 7.23& 8.73& 7.21&7.22&6.41\\
					SW & 1331.26& 1341.55&1049.01 &1264.43 &967.87 &1168.18 &1754.22 &1231.54 &873.62&2623.95\\
					EW &77067.87 &76935.58 &28039.73 &70368.97 &32594.81 &47802.08 &37289.74 &30594.15 &53100.26&54022.60\\
					EH & 1026.29&1026.36 &743.35 &1524.87 & 1082.44&689.85 &582.66 &519.06 &477.99&629.88\\
					\midrule
					Dataset  & N-BEATS & WaveNet & Transformer&Prophet & MSS$^{\bullet}$& FEDformer$^{\bullet}$& NetAtt$^{\bullet}$&  Pyraformer$^{\bullet}$ & PFSD$^{\bullet}$ & Ours\\
					\midrule
					COVID & 186.54& 1135.41& 479.96& 650.01& -& -& -& -& -& \textcolor{red}{\textbf{11.32}}\\
					Carparts & 1.11&0.74 &0.74 & 0.81& -& -&- & -& -&0.75 \\
					F-M& 2812.97& 2779.48&5098.91 &12815.47& -& -&- &- &- & \textcolor{red}{\textbf{693.75}}\\
					TW & \textcolor{red}{\textbf{1.44}}& 1.61& 1.94& 1.55& -&- &- &- &- & 1.49\\
					TH &0.02 &0.03 &0.02 &0.03& -& -&- & -&- & \textcolor{red}{\textbf{0.01}}\\
					Rideshare &6.23 &3.51 &7.17 &7.61& -& -& -& -&- & \textcolor{red}{\textbf{1.04}}\\
					Hospital & 24.18& 23.38& 40.48&20.64& -&- &- &- &- &\textcolor{red}{\textbf{20.12}} \\
					TR &11.03 &9.07 &9.01 & 9.03& -&- &- &- &- &\textcolor{red}{\textbf{6.91 }}\\
					NN5 W & \textcolor{red}{\textbf{17.35}}& 24.16& 24.02&19.60& -& -& -& -& -& 18.50\\
					NN5 D &6.47 &5.75 &5.92 &5.23& -&- & -& -& -&5.80 \\
					WTW &2820.62 &2719.37 &3815.38 &5716.74&- &- &- & -&- & \textcolor{red}{\textbf{2197.50}}\\
					WTD & -& -& -& 1130.82& -& -& -& -&- & \textcolor{red}{\textbf{465.71}}\\
					S10 & 6.62& -& 7.23& 6.94& 5.71& 6.91&7.97& 7.18&5.28 & \textcolor{red}{\textbf{0.07}}\\
					SW & 1307.78& 2569.26& 693.84& 1639.32& 972.45&\textcolor{red}{\textbf{609.94}} & 2493.06&672.54 &776.15 & 871.07\\
					EW & 35576.83& 63916.89& 78894.67& 34750.78&- & -& -& -&- &\textcolor{red}{\textbf{23086.68}} \\
					EH & 510.91& 489.91&514.68 & 630.57& -& -& -&- & -& \textcolor{red}{\textbf{301.24}}\\
					\bottomrule
			\end{tabular}}
		\end{table*}
		
		\subsection{Bilateral Atrous Algorithm (BAA)}
		After the processing of feature transformation module, temporal data in every sliding window $\delta_i$ is given relative global features. Assume the transformed information contained in segmented sequences $\delta_i$ is $\textbf{Y}_{e,i}$, we apply \textsc{BatchNorm} and \textsc{ReLu} to acquire nonlinear outputs, which can defined as:
		\begin{equation}
			\textbf{Y}_{b,i} = \textsc{ReLu}(\textsc{BatchNorm}_{\gamma, \beta}(\textbf{Y}_{e,i}))
		\end{equation}
		where $\gamma$ and $ \beta$ are learnable parameters vectors and their size are numbers of input channels. Due to the particularity of improved fuzzification of time series, the allocation of features may reduce the expressive power of neural network, so it is necessary that \textsc{BatchNorm} and \textsc{ReLu} are utilized for feature selection. However, the newly reconstructed features may be too dense and superfluous to some extent. It may lead to a need to reduce unnecessary calculation and give the proposed model bigger receptive field to further improve the ability of capturing features at a longer range. Therefore, we introduce the bilateral Atrous algorithm to build more comprehensive information modeling capabilities and greatly reduce the amount of computation needed to obtain global features. Assume processed information within windows $\delta_i$ as $\mathbf{P}(\delta_l,\varsigma_i^{'},\delta_r)$ in which $\delta_l$ and $\delta_r$ represent feature vectors lying on two sides of reconstructed elements $\varsigma^{'}_i$ and corresponding filters in Atrous algorithm as $\mathbb{F}$. For $\mathbf{P}(\delta_l,\varsigma_i^{'},\delta_r)$, the bilateral Atrous algorithm ignores every $\varsigma_i^{'}$ and slide on both sides of $\varsigma_i^{'}$ for feature acquisition to preserve the original information to the maximum extent and further process the information that is assigned to reduce the amount of computation and extract features. The process of bilateral Atrous algorithm can be defined as:
		\begin{equation}
			\begin{split}
				\textbf{Y}_{{b,i}^{'}} = \mathbf{P}(\sum_{1}^{K}\delta_l[0+s\cdot k]\circ f[k],\varsigma_i^{'},\sum_{1}^{K}\delta_r[b+s\cdot k]\circ f[k])
			\end{split}
		\end{equation}
		where $b$ represents the position of start of $\delta_r$ and $\circ$ denotes operation of dilated convolution. The bilateral Atrous algorithm preserves reconstructed elements and simplifies information, avoiding redundant calculations and makes modeling global features easier. The processed information is supposed to go through a \textsc{BatchNorm} layer before entering the next part of the model, which can be defined as:
		\begin{equation}
			\textbf{Y}_i = \textsc{BatchNorm}(\textbf{Y}_{{b,i}^{'}})
		\end{equation}
		where $\textbf{Y}_i$ represents the final output of information $\textbf{Y}_{{b,i}^{'}}$ which goes through \textsc{BatchNorm} layer.
		
\begin{figure}
	\centering 
	\includegraphics[scale=0.3]{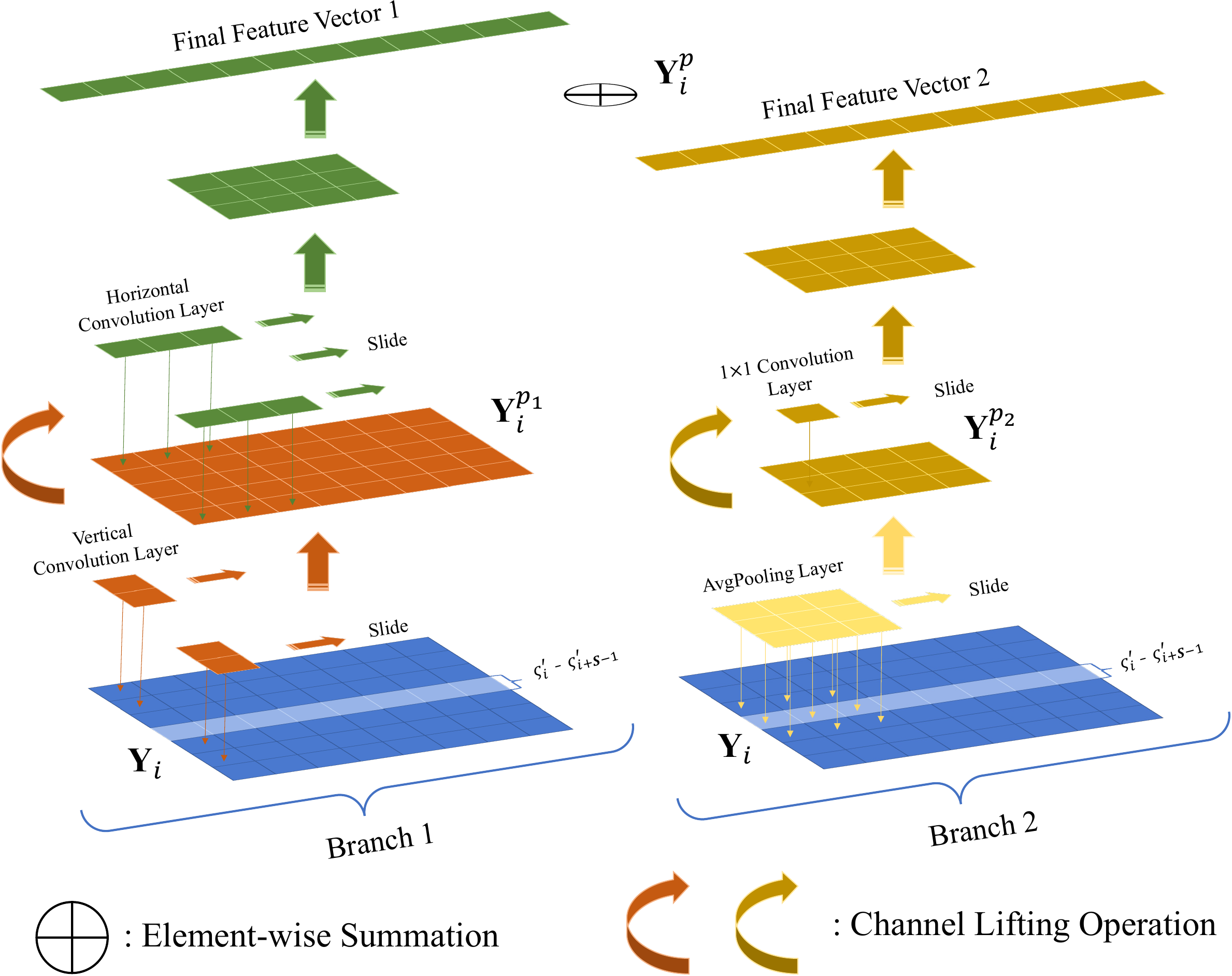}
	\caption{Details of the Partially  Asymmetric Convolutional Architecture}
	\label{3}
\end{figure}

\begin{figure*}
	\centering
	\subfigure[\qquad \qquad Comparison of MAE Among Models Without $^\bullet$]{
		\begin{minipage}[t]{0.485\linewidth}
			\centering
			\includegraphics[scale=0.34]{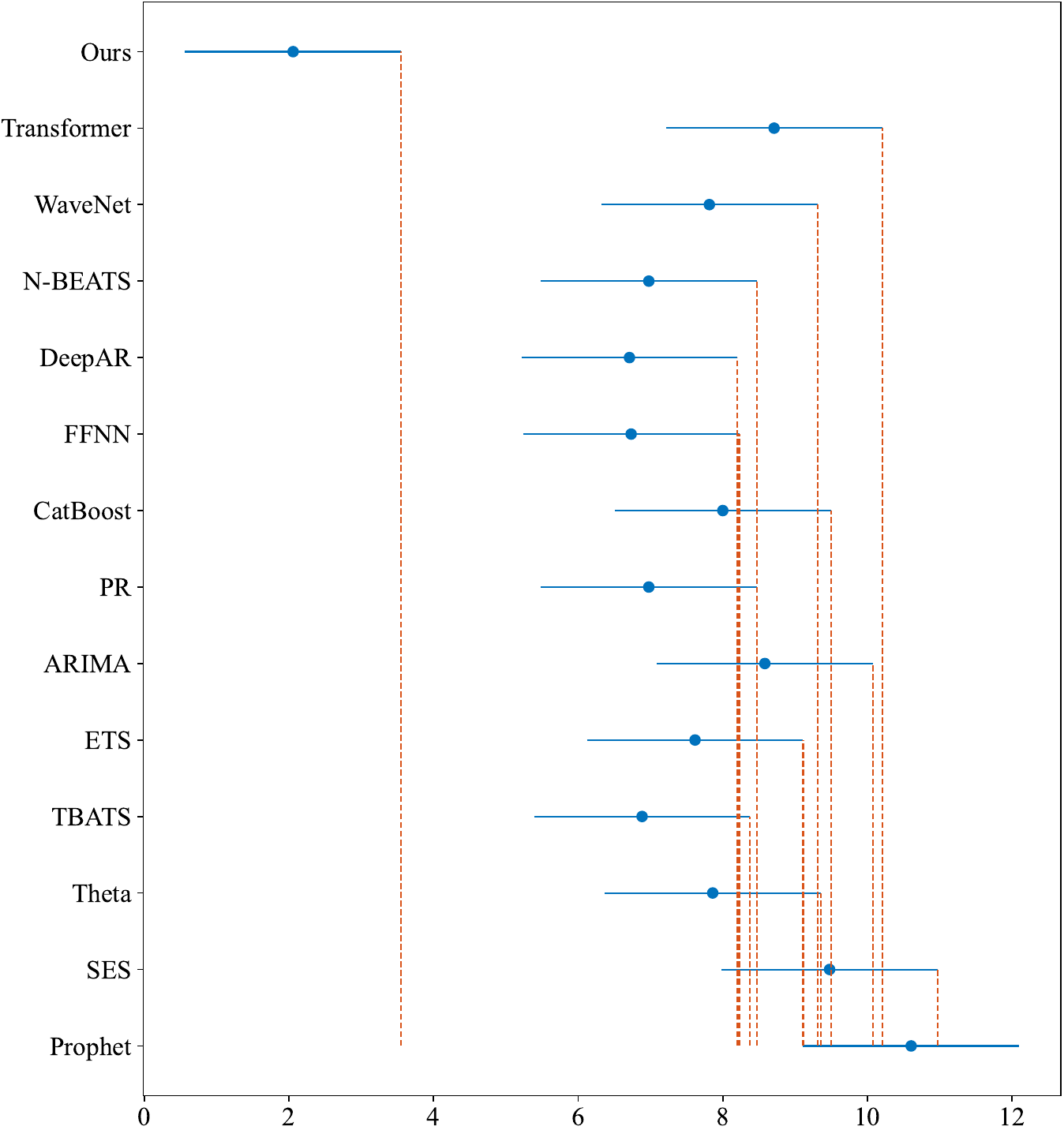}\\
		\end{minipage}%
	}%
	\subfigure[\qquad \qquad Comparison of RMSE Among Models Without $^\bullet$]{
		\begin{minipage}[t]{0.51\linewidth}
			\centering
			\includegraphics[scale=0.34]{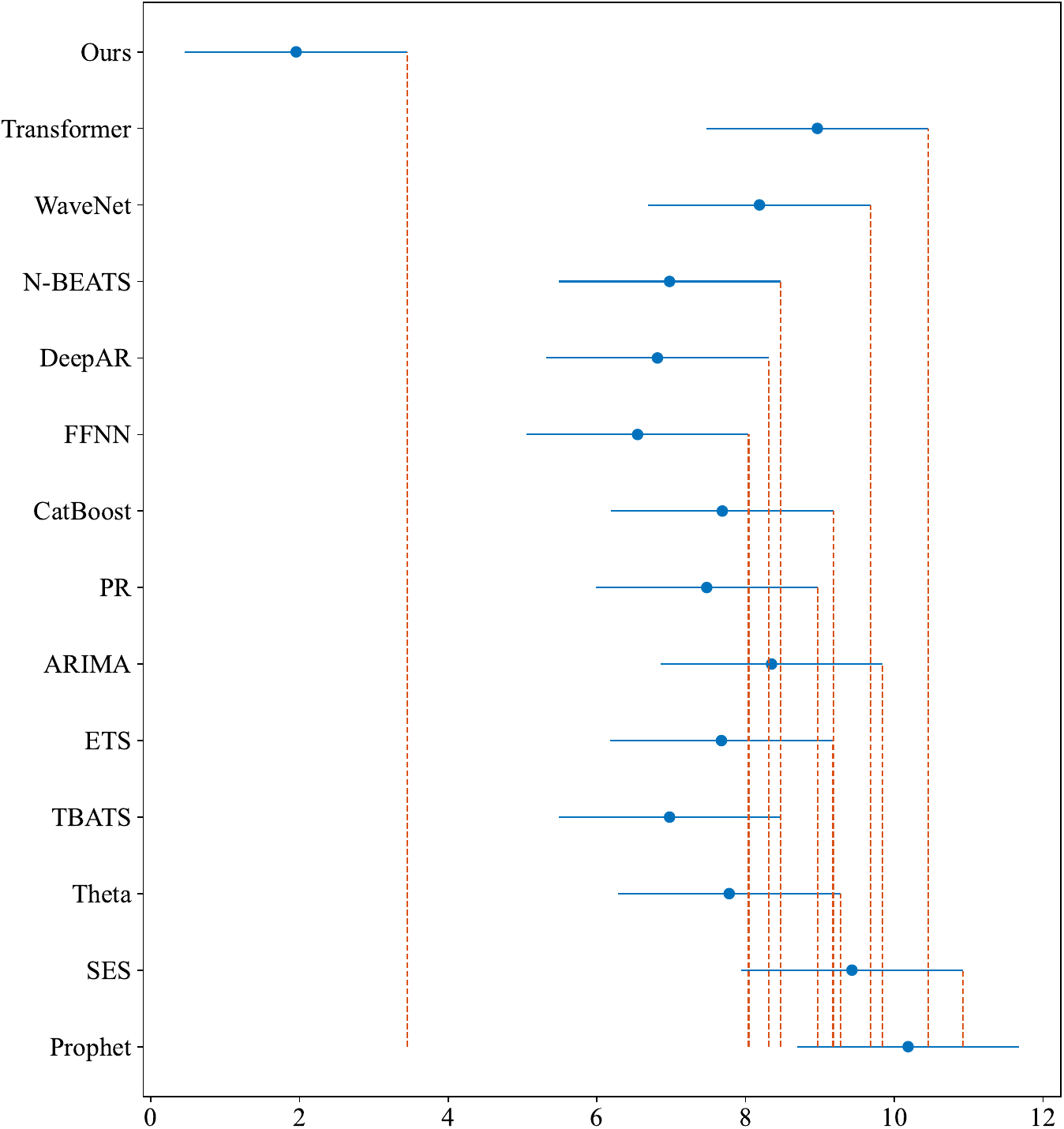}\\
		\end{minipage}%
	}%
	
	\subfigure[\qquad \qquad Comparison of MAE Among Models With $^\bullet$]{
		\begin{minipage}[t]{0.495\linewidth}
			\centering
			\includegraphics[scale=0.34]{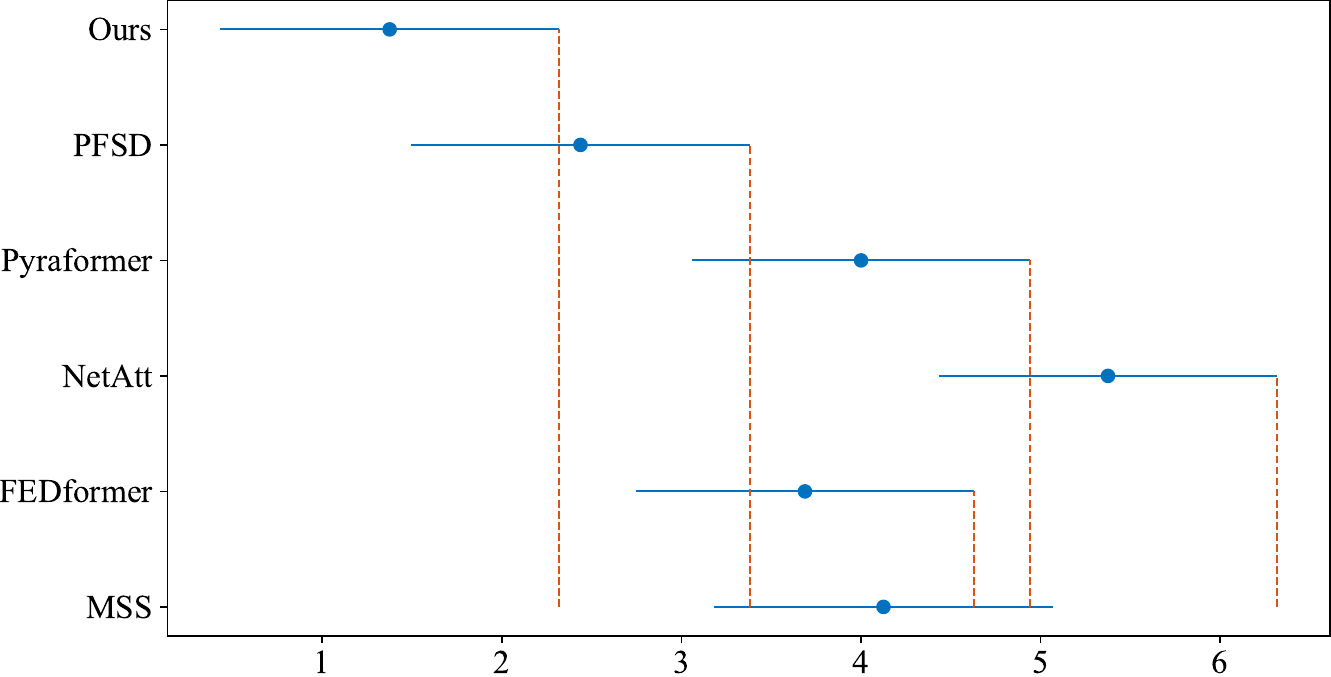}\\
		\end{minipage}%
	}%
	\subfigure[\qquad \qquad Comparison of RMSE Among Models With $^\bullet$]{
		\begin{minipage}[t]{0.49\linewidth}
			\centering
			\includegraphics[scale=0.34]{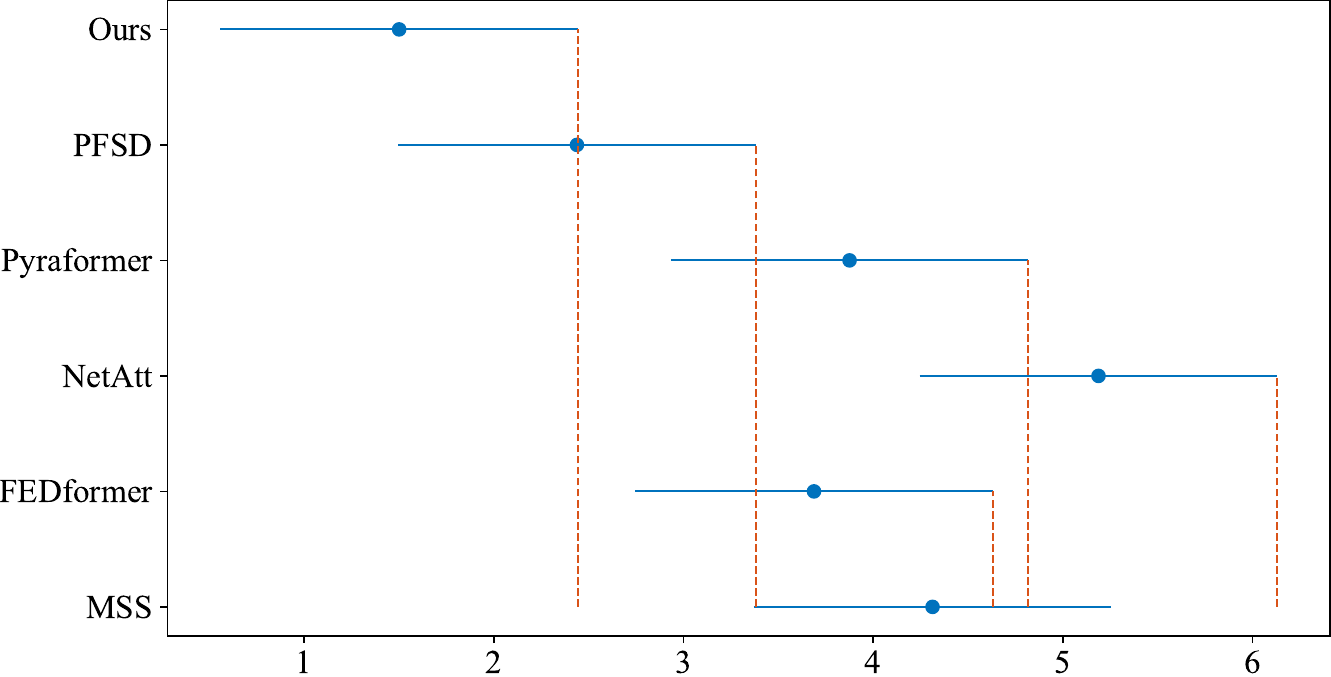}\\
		\end{minipage}%
	}%
	
	\caption{Friedman Test Figure: MAE and RMSE Comparison Among Models From the View of Nemenyi Test.}
	\label{fig6}
\end{figure*}
\subsection{Partially Asymmetric Convolutional Architecture}
Asymmetric convolution is widely utilized in the field of computer vision which reduces demand of computation resources and provides a better performance. And specifically, asymmetric convolution is a decomposition of regular convolution in which there are vertical and horizontal filters with the same length.  As a contrast, for the processed information $\textbf{Y} \in \mathbb{R}^{V^{'} \times H^{'}\times F}$, filters in partially asymmetric convolutional architecture are with variable lengths on two different directions to enable the proposed model to obtain feature from two different perspectives, namely the view of sub-windows of original sliding windows and different ranges at allocated global information, which may helps the proposed model to capture potential mutual relationships among elements. The partially asymmetric convolutional architecture which is illustrated in Figure \ref{3} consist of $K$ combinations of horizontal and vertical filters and the confirmation of value of $K$ depends on size of feature map which is mainly influenced by scale of allocated global information. Assume horizontal and vertical filters as $f_V \in \mathbb{R}^{V \times 1}$ and $f_H \in \mathbb{R}^{1 \times H}$, then the process of further extraction of information $\textbf{Y}$ can be defined as:
\begin{equation}
	\textbf{Y}^{p_1}_i = [(\textbf{Y}_i\otimes f_V)\otimes f_H]_{\times K}, \textbf{Y}^{p_1}_i \in \mathbb{R}^{V^{''} \times H^{''}\times \eta F}
\end{equation}
where $\otimes$ denotes the operation of partially asymmetric convolution and $\eta$ is the growth rate of channel numbers. If $V = H$, then the partially asymmetric convolution degenerates into the form of original asymmetric one. Generally speaking, the situation of $V = H$ is relatively rare, the sizes of the two filters are not the same in most of the time, which allows the model to build sub-windows within the original sliding window to obtain potential associations between  elements in smaller ranges, and the model also has the ability to model global information at different distances. Besides, similar to lots of effective convolutional neural network \cite{DBLP:conf/cvpr/Ding0MHD021,DBLP:journals/pami/GaoCZZYT21}, we construct residual-like branch within the proposed model to preserve original information from last combination of two kinds of filters and reduce complexity and possibility of overfitting. In detail, another branch consist of an average pooling layer with the same kernel size of regular convolution which is corresponding to two filters from the partially asymmetric convolution, $1\times1$ convolutional layer and \textsc{BatchNorm} layer which can be defined as:
\begin{equation}
	\textbf{Y}^{p_2}_i = \textsc{BatchNorm}(f_{1\times1} \diamond \textsc{AvgP}(\textbf{Y}_i)), \textbf{Y}^{p_2}_i \in \mathbb{R}^{V^{''} \times H^{''}\times G}
\end{equation}
where $\textsc{AvgP}$ represents average pooling layer and $\diamond$ denotes operation of $1\times1$ convolution. Then, with respect to the different information from two branches in the network, the fusion of it can be given as:
\begin{equation}
	\textbf{Y}^{p}_i =  \textbf{Y}^{p_1}_i \oplus \textbf{Y}^{p_2}_i
\end{equation}
where $\oplus$ represents element-wise summation of $\textbf{Y}_{p_1}$ and $\textbf{Y}_{p_2}$. And $\textbf{Y}_{p}$ is supposed to go through a bilinear layer to produce corresponding output, which can be defined as:
\begin{equation}
	\hat{\textbf{Y}}_i = \textsc{Blinear}(\textbf{Y}^{p}_i )
\end{equation}
where $	\hat{\textbf{Y}}_i$ is the final prediction made by the proposed model and $ \textsc{Blinear}$ represents two stacked linear layers. And due to the particularity of difference, there is a need to restore all of the elements within predictions to their homologous values in unknown part of time series.

\begin{figure*}
	\centering
	\subfigure[NN5 Daily Dataset]{
		\begin{minipage}[t]{0.23\linewidth}
			\centering
			\includegraphics[width=1.63in]{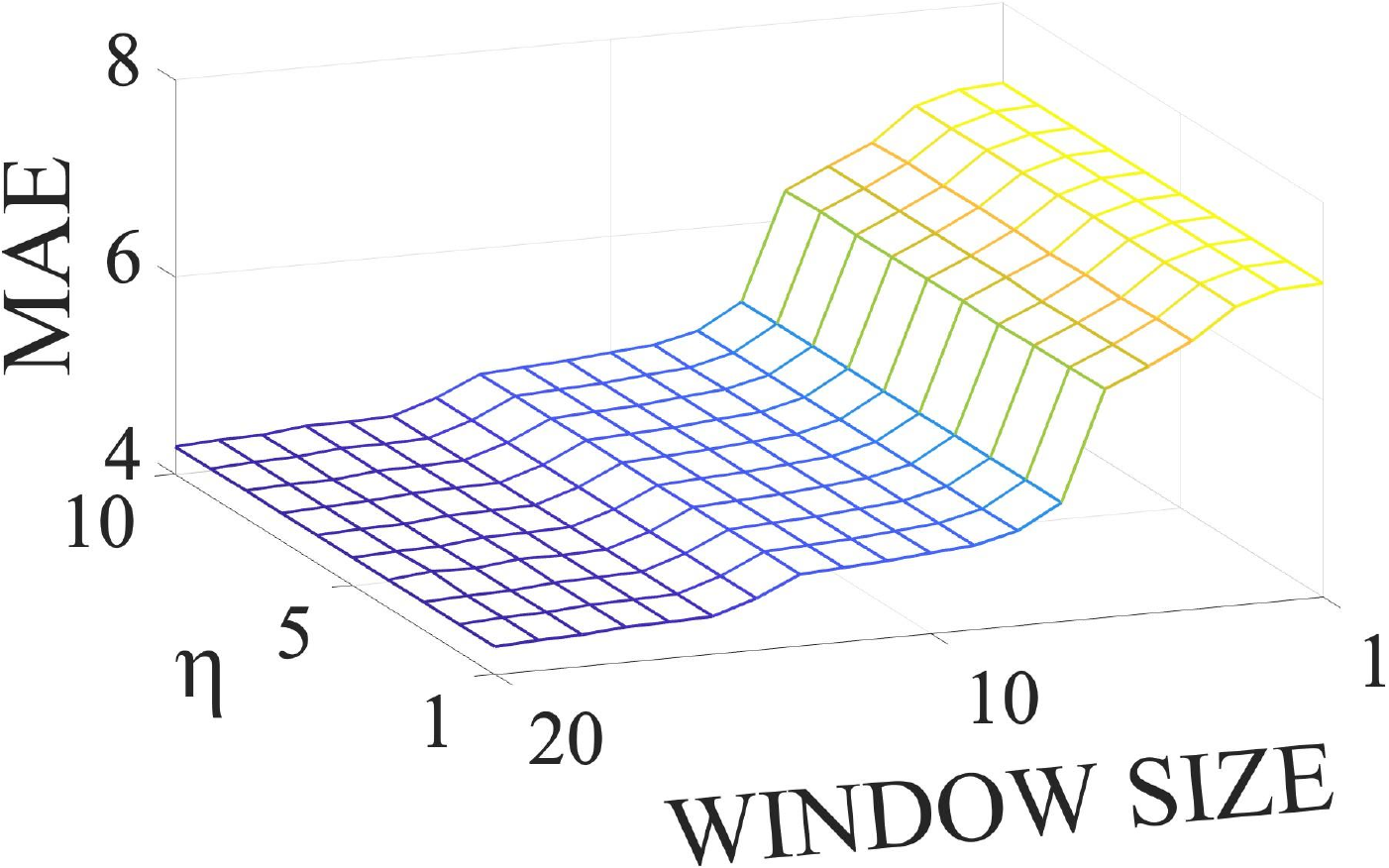}\\
			\vspace{0.6em}
			\includegraphics[width=1.47in]{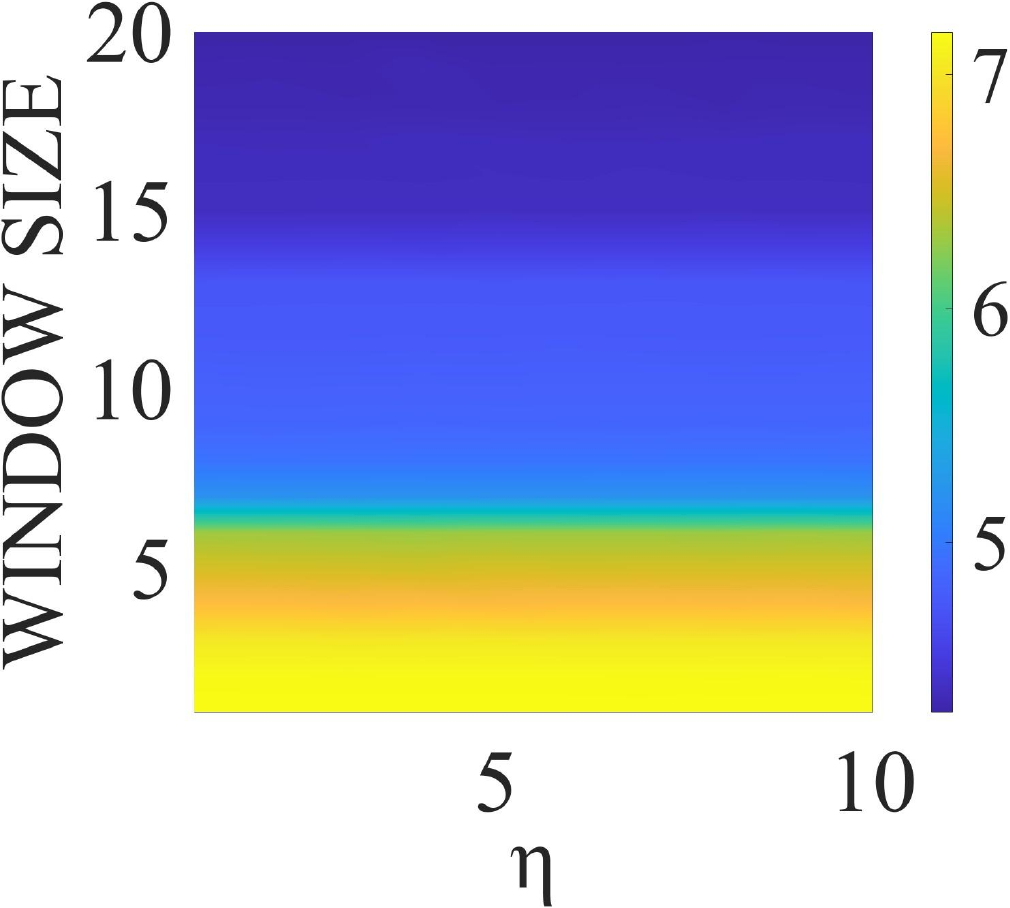}\\
			\vspace{0.5em}
		\end{minipage}%
	}%
	\subfigure[KDD Cup Dataset]{
		\begin{minipage}[t]{0.23\linewidth}
			\centering
			\includegraphics[width=1.63in]{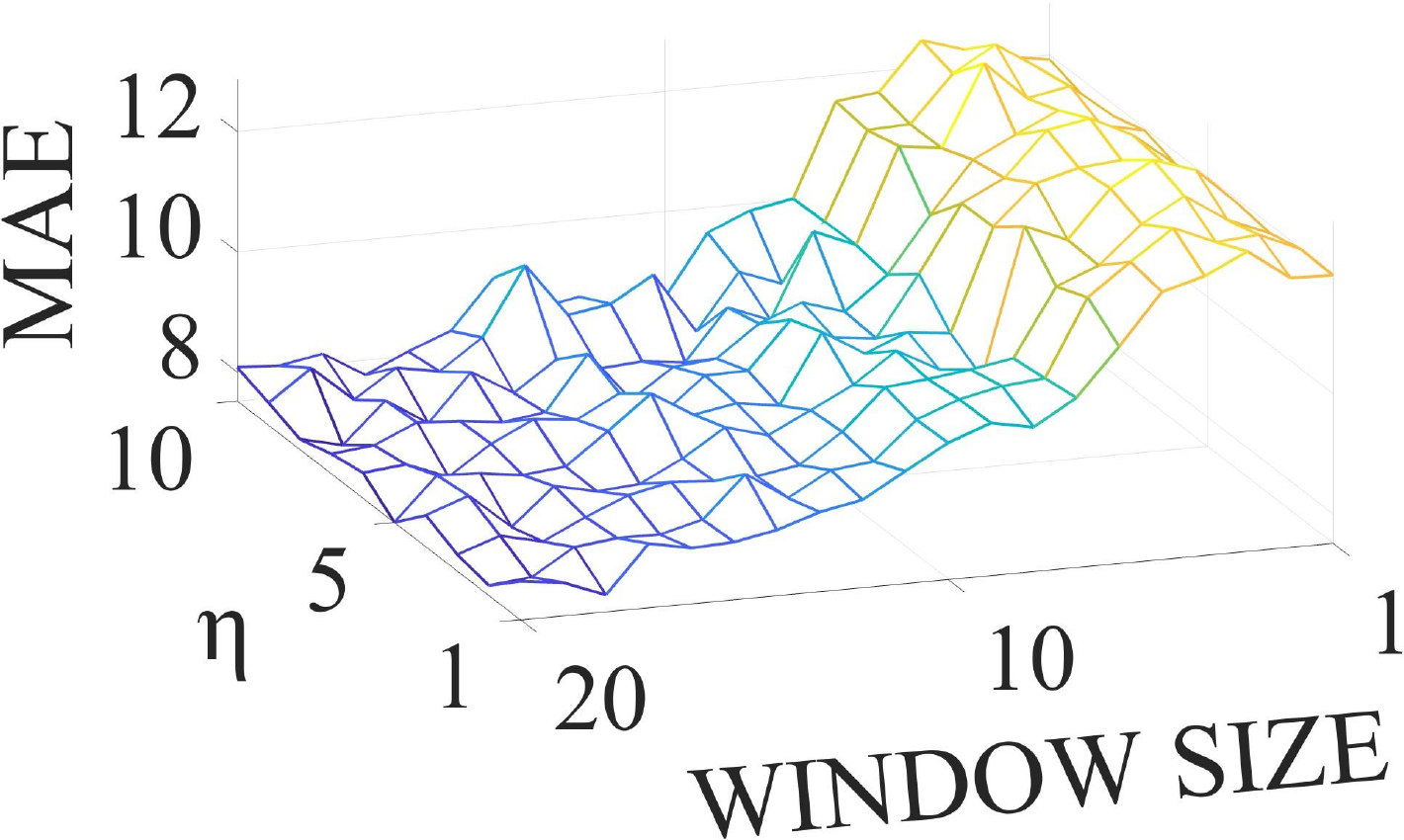}\\
			\vspace{0.65em}
			\includegraphics[width=1.6in]{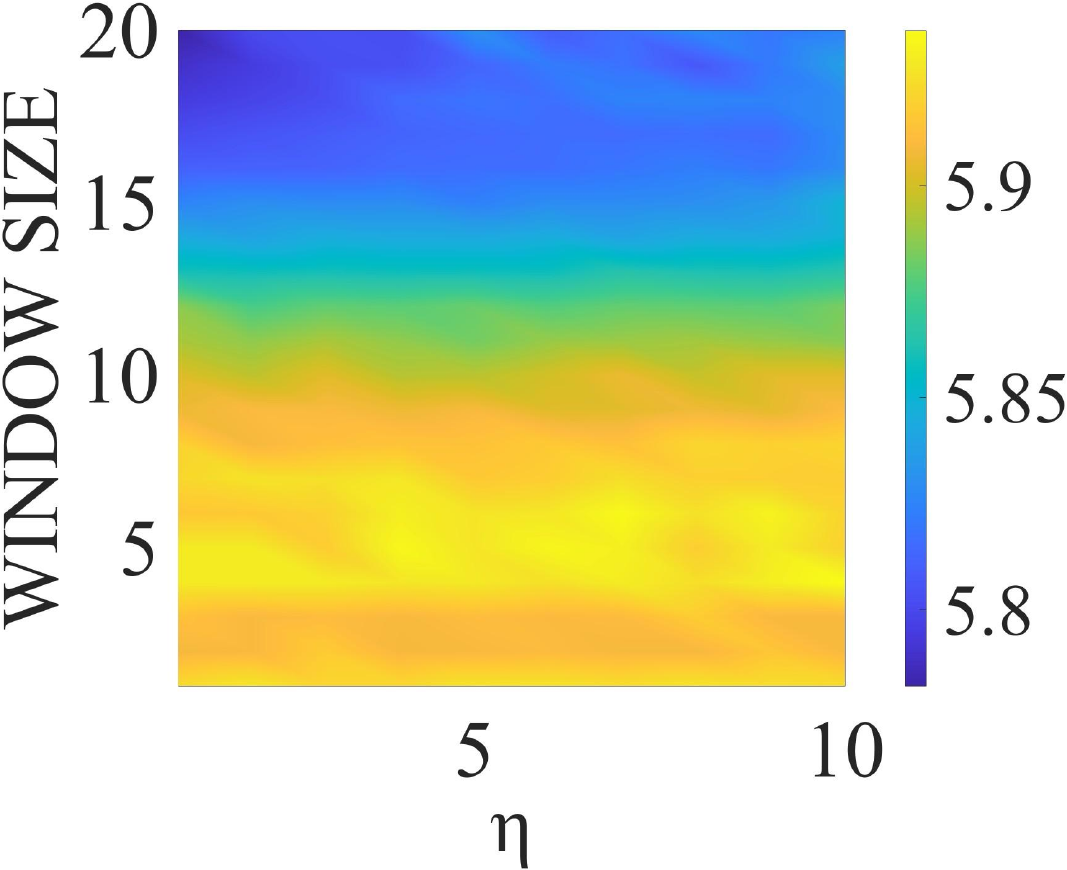}\\
			\vspace{0.5em}
		\end{minipage}%
	}%
	\subfigure[Carparts Dataset]{
		\begin{minipage}[t]{0.23\linewidth}
			\centering
			\includegraphics[width=1.7in]{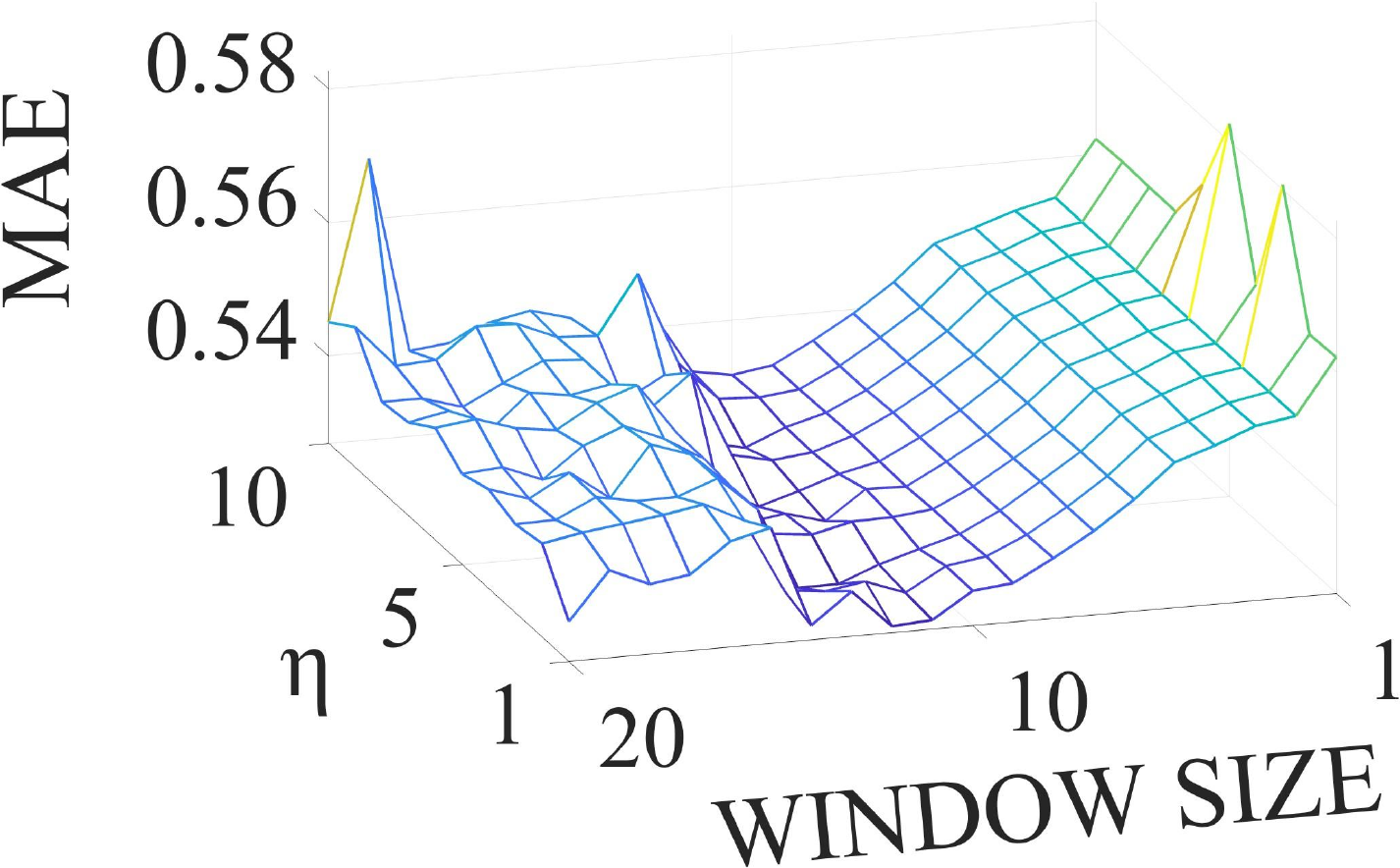}\\
			\vspace{0.8em}
			\includegraphics[width=1.63in]{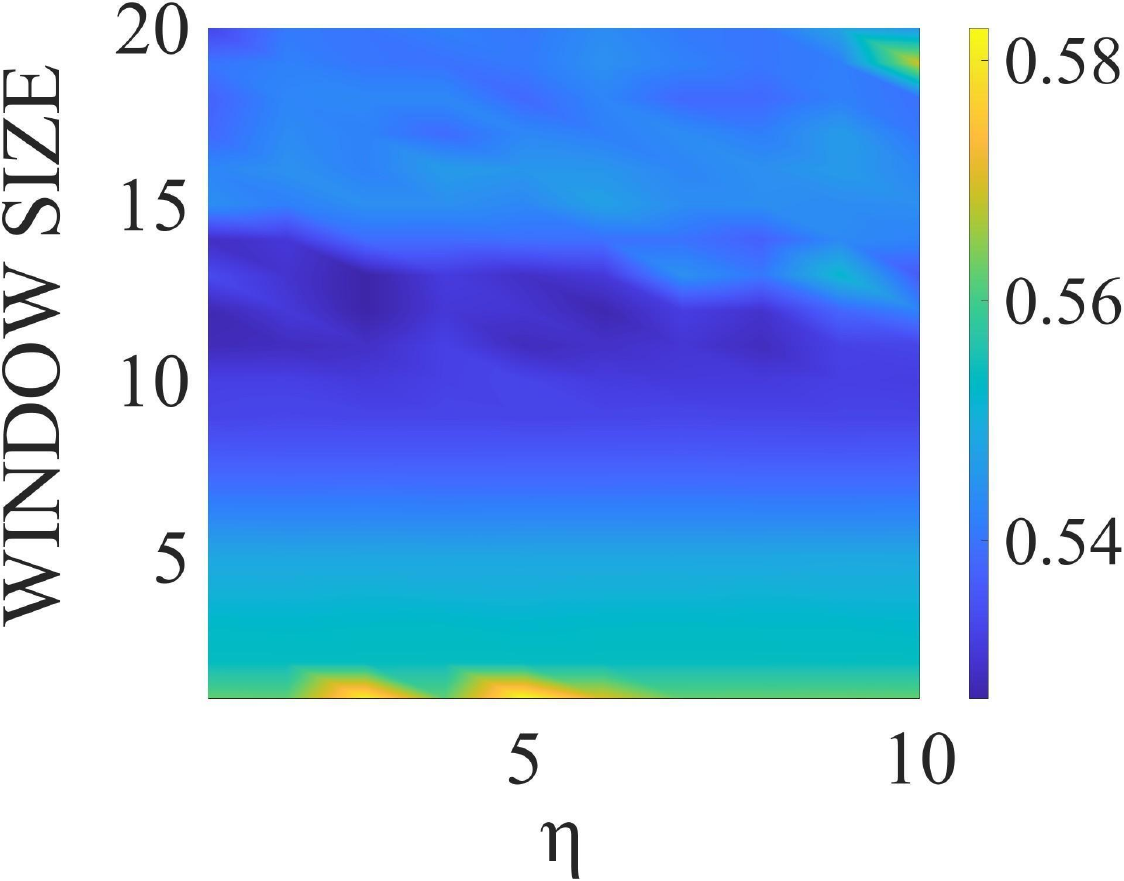}\\
			\vspace{0.5em}
		\end{minipage}%
	}%
	\subfigure[US Births Dataset]{
		\begin{minipage}[t]{0.23\linewidth}
			\centering
			\includegraphics[width=1.68in]{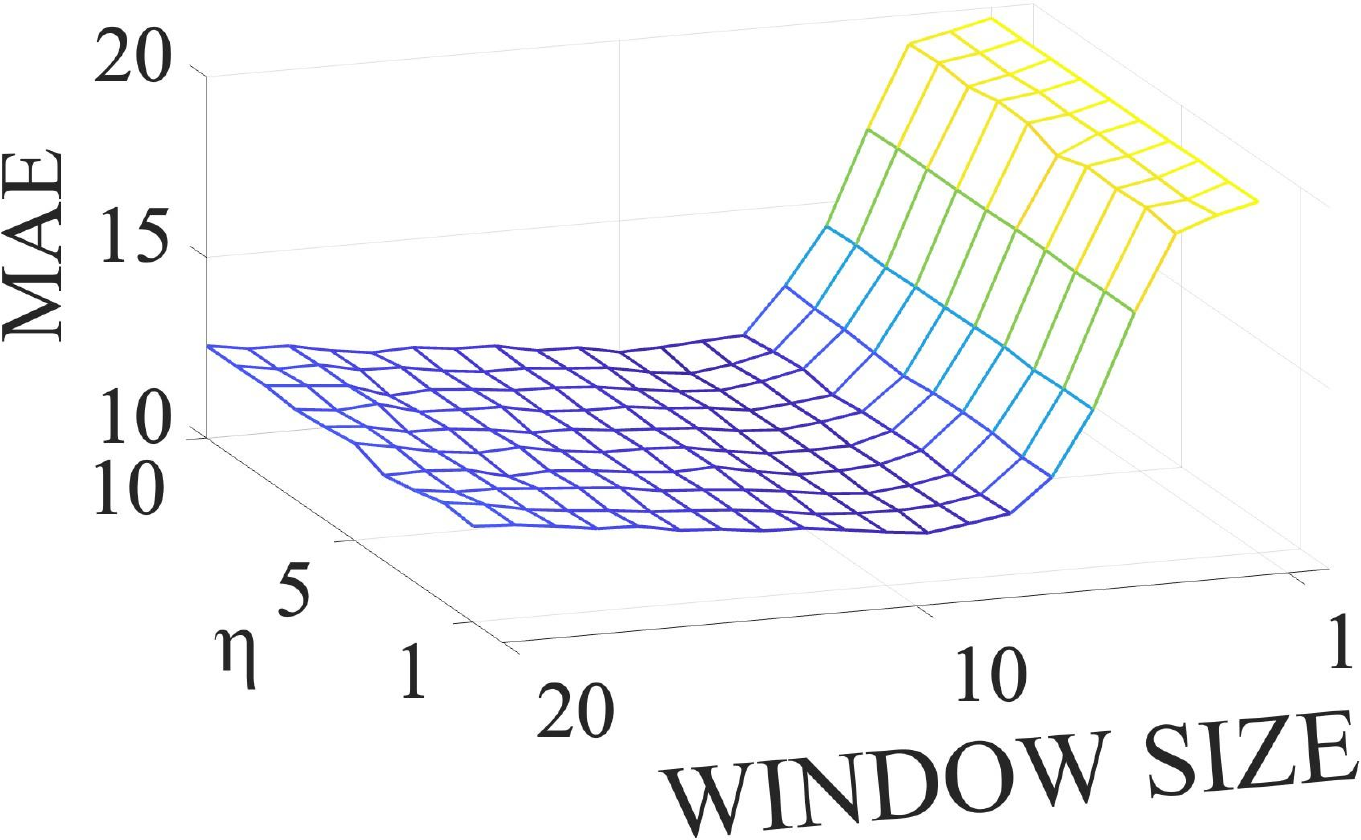}\\
			\vspace{0.8em}
			\includegraphics[width=1.53in]{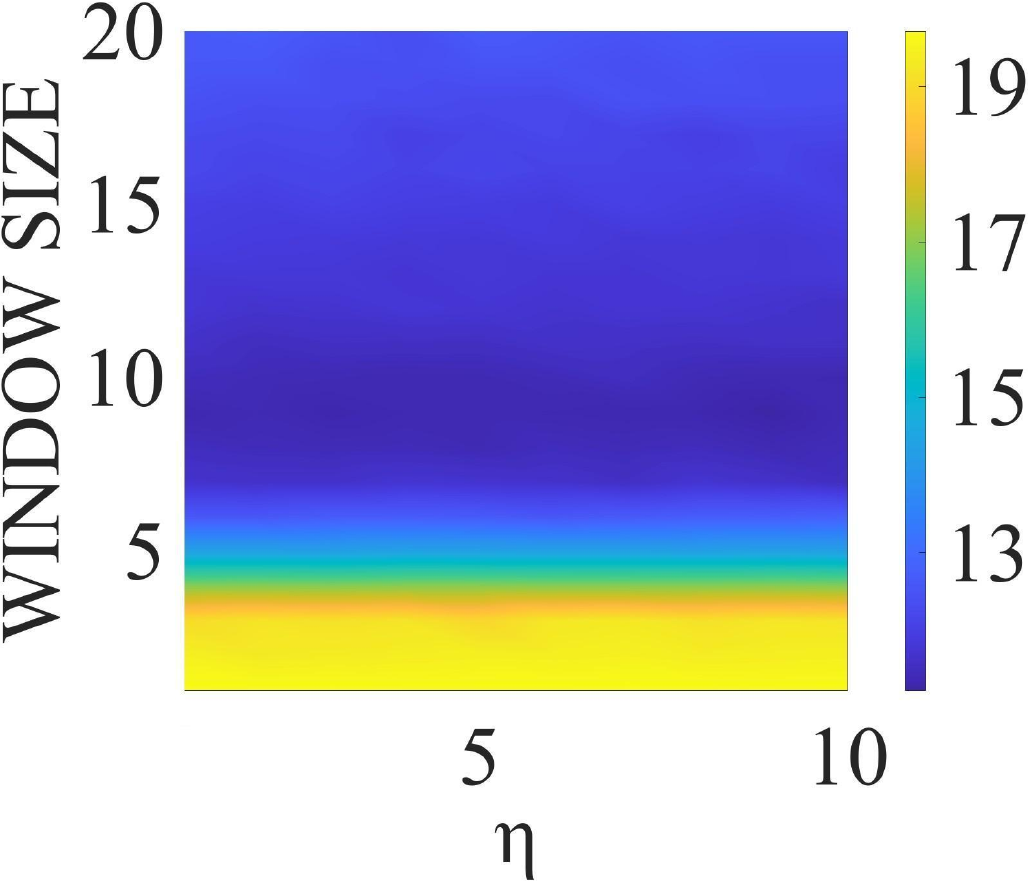}\\
			\vspace{0.5em}
		\end{minipage}%
	}%
	
	\caption{MAE Variations When Parameter $\eta$ and Window Size Vary}
	\label{fig1}
\end{figure*}

\section{Experimental evaluation}
\textbf{Datasets description}\quad In order to ensure that the experimental results can adequately measure the predictive ability of each model in different domains of time series data, we make experiments on 43 datasets whose fields include Birth rate, Weather, Traffic, Energy, Tourism, Banking and so on \cite{godahewa2021monash}. In detail, the temporal datasets cover different kinds of frequencies, such as yearly (Y), quarterly (Q) and monthly (M), and correspond to certain forecast horizons. Besides, some names of datasets are abbreviated which are listed in Table \ref{abbreviation}.

\textbf{Baselines}\quad We compare the performance of the proposed method with  modern models which represent the state-of-the-art standard of the field of time series forecasting to fully illustrate the efficiency and effectiveness of the proposed model. Specifically, we include eighteen competitive baselines: (Dynamic
Harmonic Regression-)ARIMA \cite{bookBox,hyndman2018forecasting}, Trigonometric Box-Cox ARMA Trend Seasonal Model (TBATS) \cite{articleLivera}, Simple Exponential Smoothing (SES) \cite{articleHolt}, Theta \cite{articleAssimakopoulos}, Exponential Smoothing (ETS) \cite{inbookHyndman}, Pooled Regression Model (PR) \cite{DBLP:journals/jors/TraperoKF15}, CatBoost \cite{DBLP:conf/nips/ProkhorenkovaGV18}, Feed-Forward Neural Network (FFNN) \cite{Goodfellow-et-al-2016}, DeepAR \cite{articleFlunkert},  Informer \cite{zhou2021informer}, Transformer \cite{DBLP:conf/nips/VaswaniSPUJGKP17}, FEDformer$^{\bullet}$ \cite{zhou2022fedformer}, Pyraformer$^{\bullet}$ \cite{DBLP:conf/iclr/LiuYLLLLD22}, N-BEATS \cite{DBLP:conf/iclr/OreshkinCCB20}, WaveNet \cite{borovykh2017conditional}, NetAtt$^{\bullet}$ \cite{DBLP:journals/asc/HuX22}, MSS$^{\bullet}$ \cite{articleYuntong} and PFSD$^{\bullet}$ \cite{hu2022time}. And the results of methods with $^{\bullet}$ are acquired from the paper of PFSD which are missing in some comparative datasets.  In summary, these models include both traditional and recently proposed methods, which can verify the capability of the model proposed in this paper well.

\textbf{Evaluation metrics}\quad In order to fully demonstrate model's performance, we adopt two common metrics for forecasting, namely mean absolute error (MAE) and root mean square error (RMSE). Their definitions are given as:
\begin{equation}
	MAE = \frac{1}{N}\sum_{n=1}^{N}\vert \hat{y}_n-y_n\vert
\end{equation}
\begin{equation}
	RMSE = \sqrt{\frac{1}{N}\sum_{n=1}^{N}\vert \hat{y}_n-y_n\vert^{2}}
\end{equation}
where $\hat{y}_n$ and $y_n$ represent prediction and real value  at time point $t$ respectively.

\textbf{Details of Implementation}\quad The proposed model and the whole procedures of experiments on different temporal datasets are implemented with PyTorch 2.0.0 \cite{DBLP:conf/nips/PaszkeGMLBCKLGA19}. In detail, all of the experiments are conducted on 4 NVIDIA A40 48GB GPU and 256GB RAM. Besides, to obtain experimental results on various datasets, the proposed model is trained with optimizer NAdam, scheduler ReduceLROnPlateau with eps 1e-5, threshold 1e-5, factor 0.5, patience 5, loss function L1Loss and 500 epochs.



\begin{figure}[htbp]
	\centering
	\subfigure[Prediction MAE]{
		\begin{minipage}[t]{0.45\linewidth} 
			\centering
			\includegraphics[scale=0.4]{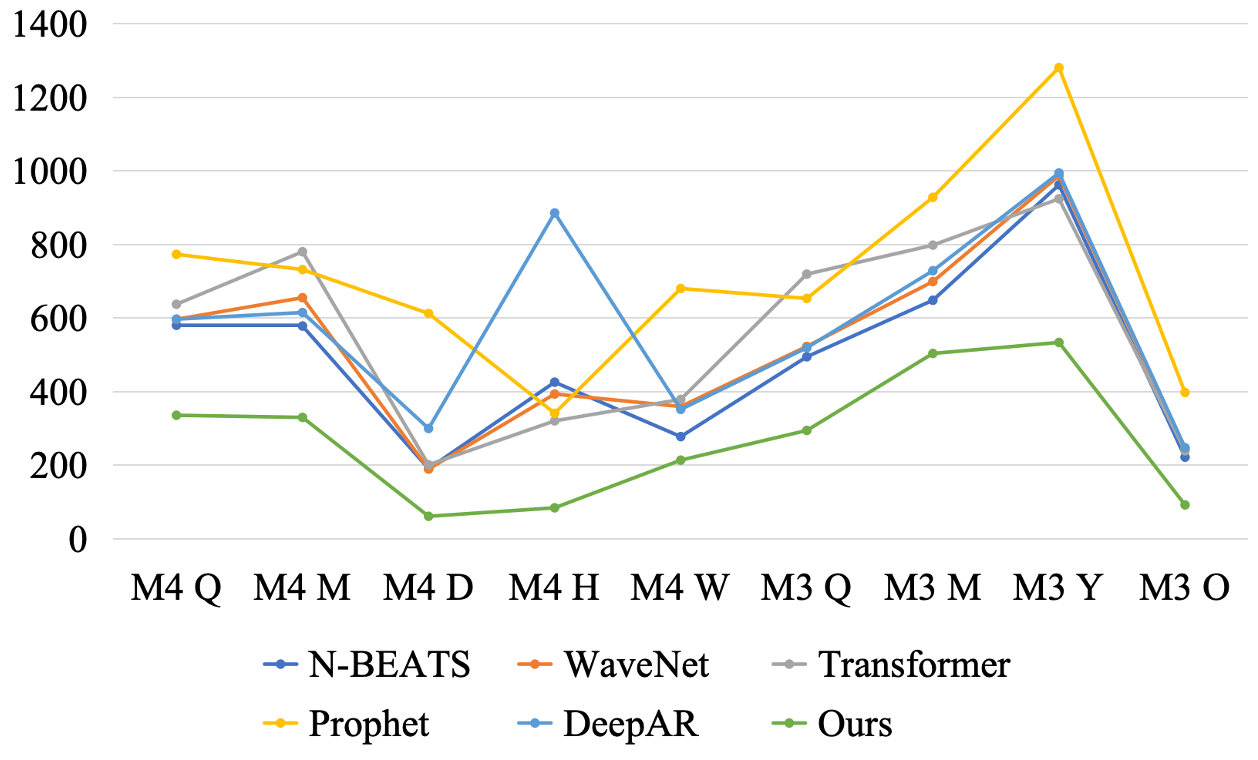}
		\end{minipage}%
	}%
	\hspace{0.05\linewidth} 
	\subfigure[Prediction RMSE]{
		\begin{minipage}[t]{0.45\linewidth}
			\centering
			\includegraphics[scale=0.42]{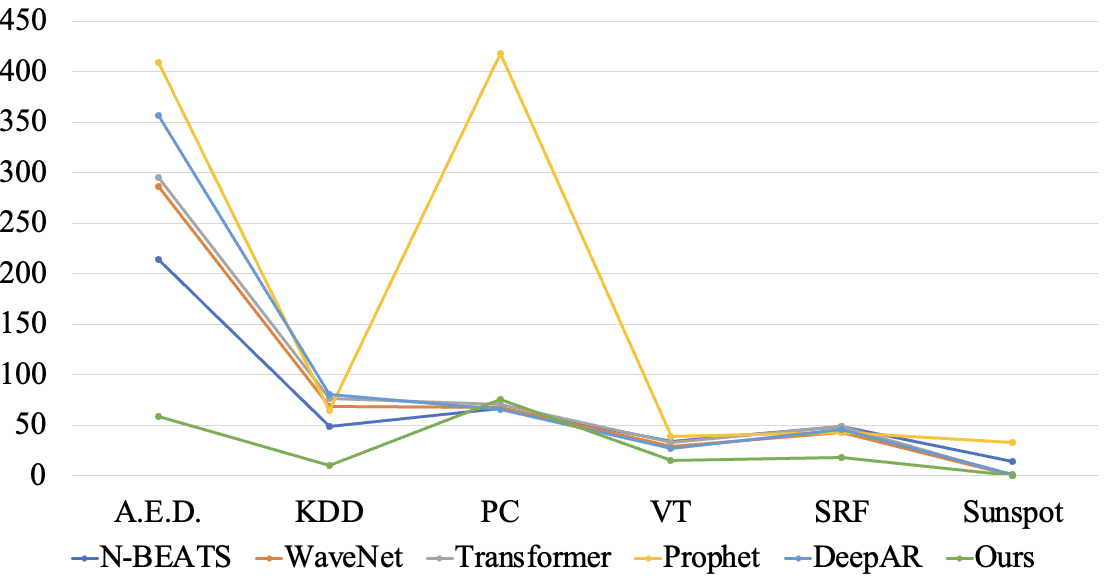}
		\end{minipage}%
	}%
	\caption{Comparison of Prediction MAE and RMSE on Various Datasets}
	\label{fig2}
\end{figure}

\subsection{Discussions on main results of experiments}
The experiments are conducted on univariate and multivariate datasets which contain data from multiple domains with separate frequencies and the experimental results are provided in Table \ref{table2}, \ref{table3}, \ref{table4} and \ref{table5}. In general, the methods proposed in this paper have achieved state-of-the-art results in most of the datasets. For univariate datasets, the method proposed in this paper achieves significantly better performance than the models proposed by other researchers in almost all of experiments. For the M series and KDD Cup competition datasets, the method proposed in this paper is better than most of traditional or modern time prediction models. Specifically, the proposed method even achieves an error reduction of more than 75$\%$ in some situations.  However, it is worth noting that the model presented in this paper does not achieve the lowest level of error in the two datasets of Sunspot and US Births, possibly because the proposed model does not memorize relatively long-term timing characteristics well. Although the corresponding global features assigned in each sliding window alleviate the problems mentioned above to some extent, the model proposed in this paper still has certain limitations in the datasets describing natural laws. For multivariable datasets, the experimental results of the model proposed in this paper are similar to those of univariable ones, and it has achieved very excellent performance in most of time series datasets in such fields as electricity power change and web traffic. But for Solar Weekly, a particular dataset which describes the laws of nature, the proposed model's predictive performance is not the most advanced. In addition, the method proposed in this paper is in a relatively backward position in multivariate NN5 competition dataset. It records the changes of the banking system, and the experimental results in this dataset show that the model proposed in this paper also has great room for improvement in the data prediction of the banking field. 

Moreover, according to the Figure \ref{fig2}, our proposed method is able to achieve very advanced prediction results on various time series data sets, and its performance is very stable compared with other modern forecasting models. In sum, experimental results demonstrate that the model proposed in this paper achieves satisfactory prediction results in both univariate and multivariate datasets. Compared with the traditional or modern methods proposed previously, the model possesses a very significant error reduction. In several datasets, the approach presented here does not achieve the most advanced performance, possibly due to its lack of ability to process long-term data features, which is an important improvement target for the later model designs. In general, the partially asymmetric convolution model proposed in this paper is capable of achieving the state-of-the-art results in most of datasets from different domains and showing very stable performance.

\subsection{Performance comparison among models from the view of Nemenyi test}
Intuitive and adequate comparison of model performance is the key to evaluate the effectiveness of models. In order to fully compare the performance of the models, we used the Nemenyi test and set $CD$ to $q_{0.05}\sqrt{\frac{k(k+1)}{6N_d}}$ where $k$ is the number of comparative algorithms and $N_d$ is the number of experimental datasets. It should be pointed out that the methods with $\bullet$, we compare them separately because the experimental results of these methods are missing on some datasets. The comparison results can be found in Figure \ref{fig6}. And it can be easily summarized that the proposed model achieves absolute leading performance no matter on univarate datasets or multivariate ones.

\subsection{Parameter Study}
In order to study the influence of window size and the factor of dimension growth rate on the prediction performance, we select two univariate datasets and two multivariate data sets for illustration and the visualizations are given in Figure \ref{fig1}. From the error variation trend of the chart, it can be concluded that our proposed method is able to obtain continuous performance gain as the time window size increases in both univariate and multivariate datasets. For example, in NN5 Daily and KDD Cup datasets, the proposed method can still observe the decrease of prediction error when the sliding window size is risen to 20. The other two datasets achieve the best prediction results when the sliding window size is around 10 to 12. In general, the proposed method can better capture the global characteristics of time series data when the amount of information obtained in each window increases, so as to achieve better prediction performance, which demonstrate the effectiveness of the proposed method in memorizing relatively long term dependencies.

\section{Conclusion}
In this paper, based on the improved fuzzy time series method, we propose a partially asymmetric convolutional architecture to realize accurate time series prediction. First, the method of transformation of temporal data in the improved fuzzification of time series allocates corresponding global information to each original time point, which gives it positional information in the view of the whole time series, so that the proposed model can better obtain the global features within each sliding window. And the process dose not require human involvement, which reduces the consumption of resources and greatly increases the reproducibility of experiments.  In addition, we also design a modified Atrous algorithm, which is used to reduce the amount of calculation required by the model and avoid the processing of redundant information to some extent while retaining the allocated temporal data characteristics. One of the most important contributions of the proposed model is that we modify the asymmetric convolution which is widely used in computer vision to enable the kind of convolution to possess variable lengths in horizontal and vertical directions. It gives the model the ability to better capture the potential relationships across different spans of elements in the existing sliding window, thus achieving more abundant feature representation. Nevertheless, the ability of obtain long-term features can be further improved by introducing recurrent design or attention mechanism, which aims to synthesizing advantages of architectures of RNN or Transformer.  In sum, the proposed model achieves state-of-the-art prediction results on most time series datasets, which strongly demonstrates the effectiveness of the proposed model and reasonableness of the modifications.

\section*{Acknowledgment}
This work is supported by National Key R$\&$D Program of China (2023YFC3502900), National Natural Science Foundation of China (granted No. 62192731) and National Key R$\&$D Program of China (2021YFF1201100). 

\bibliographystyle{elsarticle-num}
\bibliography{cite}

\begin{thebibliography}{10}
\expandafter\ifx\csname url\endcsname\relax
  \def\url#1{\texttt{#1}}\fi
\expandafter\ifx\csname urlprefix\endcsname\relax\def\urlprefix{URL }\fi
\expandafter\ifx\csname href\endcsname\relax
  \def\href#1#2{#2} \def\path#1{#1}\fi

\bibitem{DBLP:journals/pr/PangZZSZ22}
Y.~Pang, X.~Zhou, J.~Zhang, Q.~Sun, J.~Zheng, Hierarchical electricity time
  series prediction with cluster analysis and sparse penalty, Pattern Recognit.
  126 (2022) 108555.
\newblock \href {https://doi.org/10.1016/j.patcog.2022.108555}
  {\path{doi:10.1016/j.patcog.2022.108555}}.

\bibitem{DBLP:journals/eswa/GoliattY23}
L.~Goliatt, Z.~M. Yaseen, Development of a hybrid computational intelligent
  model for daily global solar radiation prediction, Expert Syst. Appl. 212
  (2023) 118295.
\newblock \href {https://doi.org/10.1016/j.eswa.2022.118295}
  {\path{doi:10.1016/j.eswa.2022.118295}}.

\bibitem{DBLP:journals/tcyb/LuRN23}
C.~Lu, C.~K. Reddy, Y.~Ning, Self-supervised graph learning with hyperbolic
  embedding for temporal health event prediction, {IEEE} Trans. Cybern. 53~(4)
  (2023) 2124--2136.
\newblock \href {https://doi.org/10.1109/TCYB.2021.3109881}
  {\path{doi:10.1109/TCYB.2021.3109881}}.

\bibitem{DBLP:journals/tai/LouLG23}
R.~Lou, Z.~Lv, M.~Guizani, Wave height prediction suitable for maritime
  transportation based on green ocean of things, {IEEE} Trans. Artif. Intell.
  4~(2) (2023) 328--337.
\newblock \href {https://doi.org/10.1109/TAI.2022.3168246}
  {\path{doi:10.1109/TAI.2022.3168246}}.

\bibitem{articleHatemi-J}
A.~Hatemi-J, R.~S. Hacker, A test for multivariate arch effects, Applied
  Economics Letters 12 (2005) 411--417.
\newblock \href {https://doi.org/10.1080/13504850500092129}
  {\path{doi:10.1080/13504850500092129}}.

\bibitem{DBLP:journals/tcyb/ChenCZWY22}
D.~Chen, L.~Chen, Y.~Zhang, B.~Wen, C.~Yang, A multiscale interactive recurrent
  network for time-series forecasting, {IEEE} Trans. Cybern. 52~(9) (2022)
  8793--8803.
\newblock \href {https://doi.org/10.1109/TCYB.2021.3055951}
  {\path{doi:10.1109/TCYB.2021.3055951}}.

\bibitem{Goodfellow-et-al-2016}
I.~Goodfellow, Y.~Bengio, A.~Courville, Deep Learning, 2016.

\bibitem{DBLP:conf/data/FabbriM18}
M.~Fabbri, G.~Moro, Dow jones trading with deep learning: The unreasonable
  effectiveness of recurrent neural networks, in: Proceedings of the 7th
  International Conference on Data Science, Technology and Applications, 2018,
  pp. 142--153.
\newblock \href {https://doi.org/10.5220/0006922101420153}
  {\path{doi:10.5220/0006922101420153}}.

\bibitem{DBLP:journals/tnn/IlhanKBK23}
F.~Ilhan, O.~Karaahmetoglu, I.~Balaban, S.~S. Kozat, Markovian {RNN:} an
  adaptive time series prediction network with hmm-based switching for
  nonstationary environments, {IEEE} Trans. Neural Networks Learn. Syst. 34~(2)
  (2023) 715--728.

\bibitem{DBLP:conf/nips/VaswaniSPUJGKP17}
A.~Vaswani, N.~Shazeer, N.~Parmar, J.~Uszkoreit, L.~Jones, A.~N. Gomez,
  L.~Kaiser, I.~Polosukhin, Attention is all you need, in: Annual Conference on
  Neural Information Processing Systems, 2017, pp. 5998--6008.

\bibitem{DBLP:conf/iclr/NieNSK23}
Y.~Nie, N.~H. Nguyen, P.~Sinthong, J.~Kalagnanam, A time series is worth 64
  words: Long-term forecasting with transformers, in: The Eleventh
  International Conference on Learning Representations, 2023.

\bibitem{DBLP:conf/nips/CunBDHHHJ89}
Y.~LeCun, B.~E. Boser, J.~S. Denker, D.~Henderson, R.~E. Howard, W.~E. Hubbard,
  L.~D. Jackel, Handwritten digit recognition with a back-propagation network,
  in: D.~S. Touretzky (Ed.), Advances in Neural Information Processing Systems
  2, {[NIPS} Conference, Denver, Colorado, USA, November 27-30, 1989], 1989,
  pp. 396--404.

\bibitem{DBLP:conf/cvpr/HeZRS16}
K.~He, X.~Zhang, S.~Ren, J.~Sun, Deep residual learning for image recognition,
  in: 2016 {IEEE} Conference on Computer Vision and Pattern Recognition, {CVPR}
  2016, Las Vegas, NV, USA, June 27-30, 2016, {IEEE} Computer Society, 2016,
  pp. 770--778.

\bibitem{DBLP:journals/tits/MaZDXW23}
C.~Ma, Y.~Zhao, G.~Dai, X.~Xu, S.~Wong, A novel {STFSA-CNN-GRU} hybrid model
  for short-term traffic speed prediction, {IEEE} Trans. Intell. Transp. Syst.
  24~(4) (2023) 3728--3737.

\bibitem{DBLP:journals/corr/abs-1803-01271}
S.~Bai, J.~Z. Kolter, V.~Koltun, An empirical evaluation of generic
  convolutional and recurrent networks for sequence modeling, CoRR
  abs/1803.01271 (2018).
\newblock \href {http://arxiv.org/abs/1803.01271} {\path{arXiv:1803.01271}}.

\bibitem{wu2023timesnet}
H.~Wu, T.~Hu, Y.~Liu, H.~Zhou, J.~Wang, M.~Long, Timesnet: Temporal
  2d-variation modeling for general time series analysis, in: International
  Conference on Learning Representations, 2023.

\bibitem{DBLP:conf/iclr/WuHLZ0L23}
H.~Wu, T.~Hu, Y.~Liu, H.~Zhou, J.~Wang, M.~Long, Timesnet: Temporal
  2d-variation modeling for general time series analysis, in: The Eleventh
  International Conference on Learning Representations, 2023.

\bibitem{Yao2024}
F.~Yao, W.~Zhao, M.~Forshaw, Y.~Song, A self-organizing interval type-2 fuzzy
  neural network for multi-step time series prediction, arXiv preprint
  arXiv:2407.08010 (2024).

\bibitem{shafi2024improved}
L.~Shafi, S.~Jain, P.~Agarwal, P.~Iqbal, A.~R. Sheergojri, An improved fuzzy
  time series forecasting model based on hesitant fuzzy sets, Journal of Fuzzy
  Extension and Applications 5~(2) (2024) 173--189.

\bibitem{phamtoan2024improving}
D.~PhamToan, N.~VoThiHang, B.~PhamThi, Improving forecasting model for fuzzy
  time series using the self-updating clustering and bi-directional long short
  term memory algorithm, Expert Systems with Applications 241 (2024) 122767.

\bibitem{Zhan2023}
T.~Zhan, Y.~He, Y.~Deng, Z.~Li, Differential convolutional fuzzy time series
  forecasting, IEEE Transactions on Fuzzy Systems 32~(3) (2023) 831--845.

\bibitem{Liu2022}
Q.~Liu, R.~Zhang, Interval type-2 fuzzy c-means forecasting model for fuzzy
  time series, Applied Soft Computing 123 (2022) 108865.

\bibitem{pinto2022interval}
A.~C.~V. Pinto, T.~E. Fernandes, P.~C. Silva, F.~G. Guimar{\~a}es, C.~Wagner,
  E.~Pestana~de Aguiar, Interval type-2 fuzzy set based time series forecasting
  using a data-driven partitioning approach, Evolving Systems 13~(5) (2022)
  703--721.

\bibitem{ashraf2024q}
S.~Ashraf, M.~S. Chohan, S.~Askar, N.~Jabbar, q-rung orthopair fuzzy time
  series forecasting technique: Prediction based decision making, AIMS
  Mathematics 9~(3) (2024) 5633--5660.

\bibitem{Chen2025}
H.~Chen, X.~Gao, Q.~Wu, An enhanced fuzzy time series forecasting model
  integrating fuzzy {C}-means clustering, the principle of justifiable
  granularity, and particle swarm optimization, Symmetry 17~(5) (2025) 753.
\newblock \href {https://doi.org/10.3390/sym17050753}
  {\path{doi:10.3390/sym17050753}}.

\bibitem{didugu2025vwfts}
G.~Didugu, M.~Gandhudi, P.~Alphonse, G.~Gangadharan, Vwfts-pso: a novel method
  for time series forecasting using variational weighted fuzzy time series and
  particle swarm optimization, International Journal of General Systems 54~(4)
  (2025) 540--559.

\bibitem{Wang2023}
S.~Wang, C.-J. Lin, A novel fuzzy time series model based on improved sparrow
  search and ceemd, Applied Soft Computing 127 (2023) 109383.

\bibitem{Singh2024}
A.~K. Singh, E.~Prasetyo, Monthly rainfall forecasting using high order singh's
  fuzzy time series, Asian Journal of Probability and Statistics 24 (2024)
  12–28.

\bibitem{Chen2024b}
X.~Chen, M.~Wu, Real-time covid-19 forecasting via a fuzzy–grey–markov
  model, Computational and Applied Mathematics 43 (2024) 25.

\bibitem{Ahmed2025}
B.~Ahmed, J.~Khan, Forecasting covid-19 active cases with a hybrid logistic
  growth and fuzzy time series model, Scientific Reports 15 (2025) 67161.

\bibitem{Kocak2024}
F.~Kocak, A.~Ozkan, Higher-order circular intuitionistic fuzzy time series
  forecasting for stock index prediction, Decision Analytics Journal 10 (2024)
  100028.

\bibitem{Alam2024}
N.~B. Alam, H.~Wahid, Stock price forecasting using fuzzy c-means and type-2
  fuzzy time series, Barekeng: Journal of Mathematics and Its Applications
  18~(2) (2024) 233–250.

\bibitem{Liu2023}
Z.~Liu, Forecasting stock prices based on multivariable fuzzy time series, AIMS
  Mathematics 8~(6) (2023) 643–662.

\bibitem{he2022new}
Y.~He, F.~Xiao, A new base function in basic probability assignment for
  conflict management, Applied Intelligence 52~(4) (2022) 4473--4487.

\bibitem{he2021conflicting}
Y.~He, F.~Xiao, Conflicting management of evidence combination from the point
  of improvement of basic probability assignment, International Journal of
  Intelligent Systems 36~(5) (2021) 1914--1942.

\bibitem{he2022mmget}
Y.~He, Y.~Deng, Mmget: a markov model for generalized evidence theory,
  Computational and Applied Mathematics 41 (2022) 1--41.

\bibitem{he2023tdqmf}
Y.~He, Y.~Deng, Tdqmf: Two-dimensional quantum mass function, Information
  Sciences 621 (2023) 749--765.

\bibitem{he2023ordinal}
Y.~He, Y.~Deng, Ordinal belief entropy, Soft Computing 27~(11) (2023)
  6973--6981.

\bibitem{he2022ordinal}
Y.~He, Y.~Deng, Ordinal fuzzy entropy, Iranian Journal of Fuzzy Systems 19~(3)
  (2022) 171--186.

\bibitem{he2024mutual}
Y.~He, Y.~Bi, L.~Li, C.-M. Pun, W.~Jiao, Z.~Jin, Mutual evidential deep
  learning for semi-supervised medical image segmentation, in: 2024 IEEE
  International Conference on Bioinformatics and Biomedicine (BIBM), IEEE,
  2024, pp. 2010--2017.

\bibitem{he2024uncertainty}
Y.~He, L.~Li, Uncertainty-aware evidential fusion-based learning for
  semi-supervised medical image segmentation, arXiv preprint arXiv:2404.06177
  (2024).

\bibitem{he2025co}
Y.~He, L.~Li, T.~Zhan, C.-M. Pun, W.~Jiao, Z.~Jin, Co-evidential fusion with
  information volume for semi-supervised medical image segmentation, Pattern
  Recognition 166 (2025) 111639.

\bibitem{he2024epl}
Y.~He, Epl: Evidential prototype learning for semi-supervised medical image
  segmentation, arXiv preprint arXiv:2404.06181 (2024).

\bibitem{he2024generalized}
Y.~He, L.~Li, T.~Zhan, W.~Jiao, C.-M. Pun, Generalized uncertainty-based
  evidential fusion with hybrid multi-head attention for weak-supervised
  temporal action localization, in: ICASSP 2024-2024 IEEE International
  Conference on Acoustics, Speech and Signal Processing (ICASSP), IEEE, 2024,
  pp. 3855--3859.

\bibitem{li2025adaptive}
L.~Li, Y.~He, C.-M. Pun, An adaptive framework for multi-view clustering
  leveraging conditional entropy optimization, in: ICASSP 2025-2025 IEEE
  International Conference on Acoustics, Speech and Signal Processing (ICASSP),
  2025.

\bibitem{li2022nndf}
L.~Li, Y.~He, L.~Li, Nndf: A new neural detection network for aspect-category
  sentiment analysis, in: International Conference on Knowledge Science,
  Engineering and Management, Springer International Publishing Cham, 2022, pp.
  339--355.

\bibitem{xu2023spatio}
T.~Xu, K.~Yan, Y.~He, S.~Gao, K.~Yang, J.~Wang, J.~Liu, Z.~Liu, Spatio-temporal
  variability analysis of vegetation dynamics in china from 2000 to 2022 based
  on leaf area index: A multi-temporal image classification perspective, Remote
  Sensing 15~(12) (2023) 2975.

\bibitem{he2021matrix}
Y.~He, L.~Li, T.~Zhan, A matrix-based distance of pythagorean fuzzy set and its
  application in medical diagnosis, arXiv preprint arXiv:2102.01538 (2021).

\bibitem{huang2025unitrans}
C.-j. Huang, Y.~He, X.~Han, W.~Jiao, Z.~Jin, L.~Wang, Unitrans: A unified
  vertical federated knowledge transfer framework for enhancing cross-hospital
  collaboration, arXiv preprint arXiv:2501.11388 (2025).

\bibitem{bi2025multi}
Y.~Bi, E.~Che, Y.~Chen, Y.~He, J.~Qu, Multi-prototype-based embedding
  refinement for medical image segmentation, in: ICASSP 2025-2025 IEEE
  International Conference on Acoustics, Speech and Signal Processing (ICASSP),
  IEEE, 2025, pp. 1--5.

\bibitem{li2024efficient}
L.~Li, Y.~He, C.-M. Pun, Efficient prototype consistency learning in
  semi-supervised medical image segmentation via joint uncertainty and data
  augmentation, in: 2024 IEEE International Conference on Bioinformatics and
  Biomedicine (BIBM), IEEE, 2024, pp. 2114--2121.

\bibitem{chen2025revisit}
X.~Chen, Z.~Tao, K.~Zhang, C.~Zhou, W.~Gu, Y.~He, M.~Zhang, X.~Cai, H.~Zhao,
  Z.~Jin, Revisit self-debugging with self-generated tests for code generation,
  arXiv preprint arXiv:2501.12793 (2025).

\bibitem{he2024residual}
Y.~He, W.~Song, L.~Li, T.~Zhan, W.~Jiao, Residual feature-reutilization
  inception network, Pattern Recognition 152 (2024) 110439.

\bibitem{li2025rethinking}
L.~Li, Z.~Jin, Y.~He, D.~Jin, H.~Duan, Z.~Tao, X.~Zhang, J.~Li, Rethinking
  regularization methods for knowledge graph completion, arXiv preprint
  arXiv:2505.23442 (2025).

\bibitem{li2025towards}
L.~Li, Z.~Jin, Y.~Zhang, D.~Jin, C.~Dou, Y.~He, X.~Zhang, H.~Zhao, Towards
  structure-aware model for multi-modal knowledge graph completion, arXiv
  preprint arXiv:2505.21973 (2025).

\bibitem{li2025multi}
L.~Li, Z.~Jin, X.~Zhang, H.~Duan, J.~Wang, Z.~Tao, H.~Zhao, X.~Zhu, Multi-view
  riemannian manifolds fusion enhancement for knowledge graph completion, IEEE
  Transactions on Knowledge and Data Engineering (2025).

\bibitem{DBLP:journals/pami/0001HLC23}
J.~Liu, Q.~Hou, Z.~Liu, M.~Cheng, Poolnet+: Exploring the potential of pooling
  for salient object detection, {IEEE} Trans. Pattern Anal. Mach. Intell.
  45~(1) (2023) 887--904.

\bibitem{DBLP:conf/ssw/OordDZSVGKSK16}
A.~van~den Oord, S.~Dieleman, H.~Zen, K.~Simonyan, O.~Vinyals, A.~Graves,
  N.~Kalchbrenner, A.~W. Senior, K.~Kavukcuoglu, Wavenet: {A} generative model
  for raw audio (2016) 125.

\bibitem{DBLP:journals/tit/Liu22}
G.~Liu, Time series forecasting via learning convolutionally low-rank models,
  {IEEE} Trans. Inf. Theory 68~(5) (2022) 3362--3380.

\bibitem{DBLP:conf/iccv/DingGDH19}
X.~Ding, Y.~Guo, G.~Ding, J.~Han, Acnet: Strengthening the kernel skeletons for
  powerful {CNN} via asymmetric convolution blocks, in: 2019 {IEEE/CVF}
  International Conference on Computer Vision, 2019, pp. 1911--1920.

\bibitem{DBLP:journals/corr/ChenPKMY14}
L.~Chen, G.~Papandreou, I.~Kokkinos, K.~Murphy, A.~L. Yuille, Semantic image
  segmentation with deep convolutional nets and fully connected crfs, in: 3rd
  International Conference on Learning Representations, {ICLR} 2015, 2015.

\bibitem{DBLP:journals/pami/ChenPKMY18}
L.~Chen, G.~Papandreou, I.~Kokkinos, K.~Murphy, A.~L. Yuille, Deeplab: Semantic
  image segmentation with deep convolutional nets, atrous convolution, and
  fully connected crfs, {IEEE} Trans. Pattern Anal. Mach. Intell. 40~(4) (2018)
  834--848.

\bibitem{DBLP:conf/iclr/LiuZL0L0022}
M.~Liu, A.~Zeng, Q.~Lai, R.~Gao, M.~Li, J.~Qin, Q.~Xu, T-wavenet: {A}
  tree-structured wavelet neural network for time series signal analysis, in:
  The Tenth International Conference on Learning Representations, {ICLR} 2022,
  Virtual Event, April 25-29, 2022, OpenReview.net, 2022.

\bibitem{DBLP:journals/iandc/Zadeh65}
L.~A. Zadeh, Fuzzy sets, Inf. Control. 8~(3) (1965) 338--353.

\bibitem{DBLP:journals/tcyb/GuoWLP21}
H.~Guo, L.~Wang, X.~Liu, W.~Pedrycz, Information granulation-based fuzzy
  clustering of time series, {IEEE} Trans. Cybern. 51~(12) (2021) 6253--6261.

\bibitem{DBLP:journals/fss/Chen96a}
S.~Chen, Forecasting enrollments based on fuzzy time series, Fuzzy Sets Syst.
  81~(3) (1996) 311--319.

\bibitem{DBLP:journals/cas/Chen02}
S.~Chen, Forecasting enrollments based on high-order fuzzy time series, Cybern.
  Syst. 33~(1) (2002) 1--16.

\bibitem{DBLP:journals/tcyb/HuWCZZP22}
J.~Hu, M.~Wu, L.~Chen, K.~Zhou, P.~Zhang, W.~Pedrycz, Weighted kernel fuzzy
  c-means-based broad learning model for time-series prediction of carbon
  efficiency in iron ore sintering process, {IEEE} Trans. Cybern. 52~(6) (2022)
  4751--4763.

\bibitem{DBLP:journals/tfs/ZhangXPDWTLL23}
D.~Zhang, Y.~Xu, Y.~Peng, C.~Du, N.~Wang, M.~Tang, L.~Lu, J.~Liu, An
  interpretable station delay prediction model based on graph community neural
  network and time-series fuzzy decision tree, {IEEE} Trans. Fuzzy Syst. 31~(2)
  (2023) 421--433.

\bibitem{DBLP:journals/tcyb/HanZXQW19}
M.~Han, S.~Zhang, M.~Xu, T.~Qiu, N.~Wang, Multivariate chaotic time series
  online prediction based on improved kernel recursive least squares algorithm,
  {IEEE} Trans. Cybern. 49~(4) (2019) 1160--1172.

\bibitem{DBLP:journals/cacm/KrizhevskySH17}
A.~Krizhevsky, I.~Sutskever, G.~E. Hinton, Imagenet classification with deep
  convolutional neural networks, Commun. {ACM} 60~(6) (2017) 84--90.

\bibitem{DBLP:conf/cvpr/SzegedyVISW16}
C.~Szegedy, V.~Vanhoucke, S.~Ioffe, J.~Shlens, Z.~Wojna, Rethinking the
  inception architecture for computer vision, in: 2016 {IEEE} Conference on
  Computer Vision and Pattern Recognition, {CVPR} 2016, Las Vegas, NV, USA,
  June 27-30, 2016, 2016, pp. 2818--2826.

\bibitem{DBLP:conf/aaai/SzegedyIVA17}
C.~Szegedy, S.~Ioffe, V.~Vanhoucke, A.~A. Alemi, Inception-v4, inception-resnet
  and the impact of residual connections on learning, in: Proceedings of the
  Thirty-First {AAAI} Conference on Artificial Intelligence, 2017, pp.
  4278--4284.

\bibitem{DBLP:conf/cvpr/XieGDTH17}
S.~Xie, R.~B. Girshick, P.~Doll{\'{a}}r, Z.~Tu, K.~He, Aggregated residual
  transformations for deep neural networks, in: 2017 {IEEE} Conference on
  Computer Vision and Pattern Recognition, 2017, pp. 5987--5995.

\bibitem{DBLP:conf/cvpr/Ding0MHD021}
X.~Ding, X.~Zhang, N.~Ma, J.~Han, G.~Ding, J.~Sun, Repvgg: Making vgg-style
  convnets great again, in: {IEEE} Conference on Computer Vision and Pattern
  Recognition, {CVPR} 2021, virtual, June 19-25, 2021, Computer Vision
  Foundation / {IEEE}, 2021, pp. 13733--13742.

\bibitem{DBLP:journals/pami/GaoCZZYT21}
S.~Gao, M.~Cheng, K.~Zhao, X.~Zhang, M.~Yang, P.~H.~S. Torr, Res2net: {A} new
  multi-scale backbone architecture, {IEEE} Trans. Pattern Anal. Mach. Intell.
  43~(2) (2021) 652--662.

\bibitem{DBLP:journals/isci/Chu95}
C.~J. Chu, \href{https://doi.org/10.1016/0020-0255(95)00021-G}{Time series
  segmentation: {A} sliding window approach}, Inf. Sci. 85~(1-3) (1995)
  147--173.
\newblock \href {https://doi.org/10.1016/0020-0255(95)00021-G}
  {\path{doi:10.1016/0020-0255(95)00021-G}}.
\newline\urlprefix\url{https://doi.org/10.1016/0020-0255(95)00021-G}

\bibitem{godahewa2021monash}
R.~Godahewa, C.~Bergmeir, G.~I. Webb, R.~J. Hyndman, P.~Montero-Manso, Monash
  time series forecasting archive, in: Neural Information Processing Systems
  Track on Datasets and Benchmarks, 2021.

\bibitem{bookBox}
B.~Box, G.~Jenkins, G.~Reinsel, G.~Ljung, Time Series Analysis: Forecasting and
  Control, Vol.~68, 2016.

\bibitem{hyndman2018forecasting}
R.~J. Hyndman, G.~Athanasopoulos, Forecasting: principles and practice, OTexts,
  2018.

\bibitem{articleLivera}
A.~De~Livera, R.~Hyndman, R.~Snyder, Forecasting time series with complex
  seasonal patterns using exponential smoothing, Journal of the American
  Statistical Association 106 (2010) 1513--1527.

\bibitem{articleHolt}
C.~Holt, Forecasting trends and seasonal by exponentially weighted averages,
  Office of Naval Research Memorandum 20 (01 1957).

\bibitem{articleAssimakopoulos}
V.~Assimakopoulos, K.~Nikolopoulos, The theta model: A decomposition approach
  to forecasting, International Journal of Forecasting 16 (2000) 521--530.

\bibitem{inbookHyndman}
R.~Hyndman, A.~Koehler, K.~Ord, R.~Snyder, Forecasting with exponential
  smoothing. the state space approach (01 2008).

\bibitem{DBLP:journals/jors/TraperoKF15}
J.~R. Trapero, N.~Kourentzes, R.~Fildes, On the identification of sales
  forecasting models in the presence of promotions, J. Oper. Res. Soc. 66~(2)
  (2015) 299--307.

\bibitem{DBLP:conf/nips/ProkhorenkovaGV18}
L.~O. Prokhorenkova, G.~Gusev, A.~Vorobev, A.~V. Dorogush, A.~Gulin, Catboost:
  unbiased boosting with categorical features, in: Advances in Neural
  Information Processing Systems 31: Annual Conference on Neural Information
  Processing Systems 2018, NeurIPS 2018, December 3-8, 2018, Montr{\'{e}}al,
  Canada, 2018, pp. 6639--6649.

\bibitem{articleFlunkert}
V.~Flunkert, D.~Salinas, J.~Gasthaus, Deepar: Probabilistic forecasting with
  autoregressive recurrent networks, International Journal of Forecasting 36
  (04 2017).

\bibitem{zhou2021informer}
H.~Zhou, S.~Zhang, J.~Peng, S.~Zhang, J.~Li, H.~Xiong, W.~Zhang, Informer:
  Beyond efficient transformer for long sequence time-series forecasting, in:
  Proceedings of the AAAI Conference on Artificial Intelligence, Vol.~35, 2021,
  pp. 11106--11115.

\bibitem{zhou2022fedformer}
T.~Zhou, Z.~Ma, Q.~Wen, X.~Wang, L.~Sun, R.~Jin, Fedformer: Frequency enhanced
  decomposed transformer for long-term series forecasting, arXiv preprint
  arXiv:2201.12740 (2022).

\bibitem{DBLP:conf/iclr/LiuYLLLLD22}
S.~Liu, H.~Yu, C.~Liao, J.~Li, W.~Lin, A.~X. Liu, S.~Dustdar, Pyraformer:
  Low-complexity pyramidal attention for long-range time series modeling and
  forecasting, in: The Tenth International Conference on Learning
  Representations, {ICLR} 2022, 2022.

\bibitem{DBLP:conf/iclr/OreshkinCCB20}
B.~N. Oreshkin, D.~Carpov, N.~Chapados, Y.~Bengio, {N-BEATS:} neural basis
  expansion analysis for interpretable time series forecasting, in: 8th
  International Conference on Learning Representations, {ICLR} 2020, Addis
  Ababa, Ethiopia, April 26-30, 2020, 2020.

\bibitem{borovykh2017conditional}
A.~Borovykh, S.~Bohte, C.~W. Oosterlee, Conditional time series forecasting
  with convolutional neural networks, arXiv preprint arXiv:1703.04691 (2017).

\bibitem{DBLP:journals/asc/HuX22}
Y.~Hu, F.~Xiao, Network self attention for forecasting time series, Appl. Soft
  Comput. 124 (2022) 109092.

\bibitem{articleYuntong}
Y.~Hu, F.~Xiao, An efficient forecasting method for time series based on
  visibility graph and multi-subgraph similarity, Chaos, Solitons \& Fractals
  160 (2022) 112243.

\bibitem{hu2022time}
Y.~Hu, F.~Xiao, Time series forecasting based on fuzzy cognitive visibility
  graph and weighted multi-subgraph similarity, IEEE Transactions on Fuzzy
  Systems (2022).

\bibitem{DBLP:conf/nips/PaszkeGMLBCKLGA19}
A.~Paszke, S.~Gross, F.~Massa, A.~Lerer, J.~Bradbury, G.~Chanan, T.~Killeen,
  Z.~Lin, N.~Gimelshein, L.~Antiga, A.~Desmaison, A.~K{\"{o}}pf, E.~Z. Yang,
  Z.~DeVito, M.~Raison, A.~Tejani, S.~Chilamkurthy, B.~Steiner, L.~Fang,
  J.~Bai, S.~Chintala, Pytorch: An imperative style, high-performance deep
  learning library, in: Annual Conference on Neural Information Processing
  Systems, 2019, pp. 8024--8035.

\end{thebibliography}

\end{document}